\documentclass[sigconf]{acmart}
\usepackage{amsmath}
\usepackage{amsthm}
\usepackage{ulem}
\usepackage{mathrsfs}
\usepackage{indentfirst}
\usepackage[ruled]{algorithm2e}
\usepackage{bm}
\usepackage{graphicx}
\usepackage{subfigure}
\usepackage{booktabs}
\usepackage{footnote}
\makesavenoteenv{tabular}
\usepackage{tabularx}
\usepackage{pifont} 
\usepackage{makecell} 
\usepackage{geometry} 
\usepackage{multirow} 
\usepackage{ulem}
\useunder{\uline}{\ul}{}
\AtBeginDocument{%
  }

\setcopyright{acmlicensed}
\copyrightyear{2026}
\acmYear{2026}
\acmDOI{XXXXXXX.XXXXXXX}
\begin{document}

\title{GenTS: A Comprehensive Benchmark Library  for Generative Time Series Models}

\author{Chenxi Wang}
\affiliation{%
  \institution{The University of Hong Kong}
  \city{Hong Kong SAR}
  \state{}
  \country{China}
}
\email{cxwang@eee.hku.hk}

\author{Xiaorong Wang}
\affiliation{%
  \institution{The University of Hong Kong}
  \city{Hong Kong SAR}
  \state{}
  \country{China}
}
\email{xrwong@connect.hku.hk}

\author{Peiyang Li}
\affiliation{%
  \institution{Fudan University}
  \city{Shanghai}
  \country{China}
}
\email{pyli22@m.fudan.edu.cn}

\author{Yi Wang}
\affiliation{%
  \institution{The University of Hong Kong}
  \city{Hong Kong SAR}
  \state{}
  \country{China}
}
\email{yiwang@eee.hku.hk}
\authornote{Corresponding author}





\renewcommand{\shortauthors}{Trovato et al.}

\begin{abstract}

  Generative models have demonstrated remarkable potential in time series analysis tasks, like synthesis, forecasting, imputation, etc. However, offering limited coverage for generative models, existing time series libraries are mainly engineered for discriminative models, with standardized workflows for specific tasks, such as optimizing Mean Squared Errors for time series forecasting. This rigid structure is fundamentally incompatible with the distinct and often complex paradigms of generative models (e.g., adversarial training, diffusion processes), which learn the underlying data distribution rather than a direct input-output mapping. To this end, we proposed \texttt{GenTS}, a comprehensive and extensible benchmark library designed for systematic assessment on generative time series models. \texttt{GenTS} features a unified data preprocessing pipeline, a collection of versatile models, and panoramic evaluation metrics. Its modular design also enables the researchers to flexibly customize beyond our built-in datasets and models. Based on \texttt{GenTS}, we conducted benchmarking experiments under diverse tasks, accordingly offering suggestions for model selection and identifying potential directions for future research. Our codes are open-source at \url{https://github.com/WillWang1113/GenTS}. The official tutorials and document are available at \url{https://willwang1113.github.io/GenTS/}.

  
  \end{abstract}





\maketitle

\section{Introduction}


Time series, defined as a sequence of data points collected temporally, are ubiquitous across multiple domains, for example, financial markets, energy systems, traffic systems, and so on. Over the past decade, the success of neural networks, coupled with the increasing availability of real-world time series data, have advanced time series analysis from traditional models towards deep learning-based models \cite{wang2024deep}.


In recent years, deep generative models have once again revolutionized time series analysis. Their excellence in learning complex data distributions, such as images \cite{ho2020denoising, brock2018large, razavi2019generating}, audios \cite{van2016wavenet}, and texts \cite{achiam2023gpt, guo2025deepseek, team2023gemini}, motivates the community to further investigate generative time series models for time series analysis, where both unconditional and conditional tasks are of interest. Unconditional models learn the data distribution from which novel but realistic time series can be sampled, primarily for tasks such as data augmentation. Meanwhile, conditional models generate time series based on given auxiliary inputs, empowering classical time series analysis, such as probabilistic forecasting, which produces future scenarios from past observations. So far, pioneering generative time series models have been developed for synthesis \cite{naiman2024utilizing,yoon2019time,desai2021timevae}, imputation \cite{tashiro2021csdi,yuan2024diffusion}, forecasting \cite{rasul2021autoregressive, liu2022non}, super-resolution \cite{wang2025a}, etc.


Despite the growing research trend, there is a lack of a comprehensive benchmarking library for generative time series models. As we will introduce in Section~\ref{chap:modeloverview}, predominantly engineered for discriminative models, the existing time series libraries either hardly integrate the generative models or focus on a single generation task, ignoring the versatility. Besides, generative models often employ diverse training methods, such as adversarial training, diffusion, and denoising, that are distinct from those of discriminative models. This makes it more difficult to integrate them into existing time-series libraries. Thus, there is still a lack of a comprehensive library for generative time series models, hindering systematic assessments of new advancements and innovations. 


\begin{table*}[htbp]
\centering
\caption{Comparison of Time Series Generation Libraries. Syn.: Synthesis (either unconditional or class label-guided), Fcst: Forecasting, Impt.: Imputation, Diff.: Diffusion, D.E.: Differential Equation, M-F: Model-Free, M-B: Model-Based, Vis: Visualization}
\label{tab:restructured_comparison}

\begin{tabularx}{\linewidth}{l *{12}{>{\centering\arraybackslash}X}}
\toprule
& \multicolumn{1}{c}{\textbf{Datasets}} 
& \multicolumn{5}{c}{\textbf{Models}} 
& \multicolumn{3}{c}{\textbf{Tasks}} 
& \multicolumn{3}{c}{\textbf{Evaluation}} \\
\cmidrule(lr){2-2} \cmidrule(lr){3-7} \cmidrule(lr){8-10} \cmidrule(lr){11-13}

& \textbf{Diversity}
& \textbf{GAN} & \textbf{VAE} & \textbf{Flow} & \textbf{Diff.} & \textbf{D. E.}
& \textbf{Syn.} & \textbf{Fcst.} & \textbf{Imp.}
& \textbf{M-F} & \textbf{M-B} & \textbf{Vis.} \\
\midrule

Synthcity \cite{qian2023synthcity} &
\textcolor{red}{\ding{55}} &
\textcolor{green}{\ding{51}} & \textcolor{green}{\ding{51}} & \textcolor{green}{\ding{51}} & \textcolor{green}{\ding{51}} & \textcolor{red}{\ding{55}} &
\textcolor{green}{\ding{51}} & \textcolor{red}{\ding{55}} & \textcolor{red}{\ding{55}} &
\textcolor{green}{\ding{51}} & \textcolor{green}{\ding{51}} & \textcolor{red}{\ding{55}} \\
\addlinespace

TSGM \cite{nikitin2024tsgm} &
\textcolor{green}{\ding{51}} &
\textcolor{green}{\ding{51}} & \textcolor{green}{\ding{51}} & \textcolor{red}{\ding{55}} & \textcolor{green}{\ding{51}} & \textcolor{red}{\ding{55}} &
\textcolor{green}{\ding{51}} & \textcolor{red}{\ding{55}} & \textcolor{red}{\ding{55}} &
\textcolor{green}{\ding{51}} & \textcolor{green}{\ding{51}} & \textcolor{green}{\ding{51}} \\
\addlinespace

TSLib \cite{wang2024deep} &
\textcolor{green}{\ding{51}} &
\textcolor{red}{\ding{55}} & \textcolor{red}{\ding{55}} & \textcolor{red}{\ding{55}} & \textcolor{red}{\ding{55}} & \textcolor{red}{\ding{55}} &
\textcolor{red}{\ding{55}} & \textcolor{green}{\ding{51}} & \textcolor{green}{\ding{51}} &
\textcolor{green}{\ding{51}} & \textcolor{red}{\ding{55}} & \textcolor{red}{\ding{55}} \\
\addlinespace

GluonTS \cite{gluonts_jmlr} &
\textcolor{green}{\ding{51}} &
\textcolor{red}{\ding{55}} & \textcolor{red}{\ding{55}} & \textcolor{red}{\ding{55}} & \textcolor{red}{\ding{55}} & \textcolor{red}{\ding{55}} &
\textcolor{red}{\ding{55}} & \textcolor{green}{\ding{51}} & \textcolor{red}{\ding{55}} &
\textcolor{green}{\ding{51}} & \textcolor{red}{\ding{55}} & \textcolor{red}{\ding{55}} \\
\addlinespace

ProbTS \cite{zhang2024probts} &
\textcolor{green}{\ding{51}} &
\textcolor{red}{\ding{55}} & \textcolor{red}{\ding{55}} & \textcolor{green}{\ding{51}} & \textcolor{green}{\ding{51}} & \textcolor{red}{\ding{55}} &
\textcolor{red}{\ding{55}} & \textcolor{green}{\ding{51}} & \textcolor{red}{\ding{55}} &
\textcolor{green}{\ding{51}} & \textcolor{red}{\ding{55}} & \textcolor{red}{\ding{55}} \\
\addlinespace

TSGBench \cite{ang2023tsgbench} &
\textcolor{green}{\ding{51}} &
\textcolor{orange}{\ding{115}} & \textcolor{orange}{\ding{115}} & \textcolor{orange}{\ding{115}} & \textcolor{orange}{\ding{115}} & \textcolor{orange}{\ding{115}} &
\textcolor{green}{\ding{51}} & \textcolor{red}{\ding{55}} & \textcolor{red}{\ding{55}} &
\textcolor{green}{\ding{51}} & \textcolor{green}{\ding{51}} & \textcolor{green}{\ding{51}} \\
\addlinespace

CTBench \cite{ang2025ctbench} &
\textcolor{red}{\ding{55}} &
\textcolor{orange}{\ding{115}} & \textcolor{orange}{\ding{115}} & \textcolor{orange}{\ding{115}} & \textcolor{orange}{\ding{115}} & \textcolor{orange}{\ding{115}} &
\textcolor{green}{\ding{51}} & \textcolor{red}{\ding{55}} & \textcolor{red}{\ding{55}} &
\textcolor{green}{\ding{51}} & \textcolor{green}{\ding{51}} & \textcolor{green}{\ding{51}} \\
\addlinespace

\textbf{GenTS (Ours)} &
\textcolor{green}{\ding{51}} &
\textcolor{green}{\ding{51}} & \textcolor{green}{\ding{51}} & \textcolor{green}{\ding{51}} & \textcolor{green}{\ding{51}} & \textcolor{green}{\ding{51}} &
\textcolor{green}{\ding{51}} & \textcolor{green}{\ding{51}} & \textcolor{green}{\ding{51}} &
\textcolor{green}{\ding{51}} & \textcolor{green}{\ding{51}} & \textcolor{green}{\ding{51}} \\

\bottomrule
\end{tabularx}
\parbox{\textwidth}{
\vspace{2mm}
\footnotesize
In this table, \textcolor{green}{\ding{51}} indicates inclusion while \textcolor{red}{\ding{55}} denotes exclusion. Primarily focusing on systematic evaluation metrics of generated time series data, TSGBench and CTBench are labeled with
\textcolor{orange}{\ding{115}} in terms of Models, since their official code repositories only provide a template base model class, without implementation on SOTA models.
}
\end{table*}

To this end, we are motivated to propose \texttt{GenTS}, a comprehensive benchmark library for time series generation. In \texttt{GenTS}, we unified the preprocessing pipeline of time series data with built-in multi-domain datasets, collected the versatile generative time series models, and constructed the panoramic performance evaluation for generated time series under different contexts. Modularly designed, \texttt{GenTS} also enables users to customize their own models and datasets, thereby further contributing to the broader scientific community. We also launched large-scale benchmarking experiments under different time series generation tasks to disclose the strengths and weaknesses of different types of generative time series models.

\textbf{Contributions:} 1) We proposed a comprehensive benchmark library, \texttt{GenTS}, designed for generative time series models. It provides standardized but also customizable workflow of data preparation, model training and performance evaluation for unconditional and conditional time series generation tasks. 2) Based on \texttt{GenTS}, we launched benchmark experiments over three common and major time series generation tasks, i.e., synthesis, forecasting and imputation, which can be referred to for other related research works. 3) Accordingly, we formed suggestions on model selections in cross-task and task-specific contexts, and provided some potential future directions for the community.


\section{Preliminaries}
In this section, we will introduce the definition of time series generation and our focused time series generation tasks in this study.
\subsection{Definition of Time Series Generation}


Let us denote a multivariate time series as $\boldsymbol{X}=(\boldsymbol{x}_1,\boldsymbol{x}_2, \cdots, \boldsymbol{x}_T)^\top \in \mathbb{R}^{T \times D}$, where $T$ is the length of time series, $D$ is the number of variables, and for $t \in \{1, \cdots, T \}$, $\boldsymbol{x}_t \in \mathbb{R}^{D}$ is the vector of observations at time step $t$. We are interested in the data distribution $p(\boldsymbol{X})$ and in practice, we have access to the time series dataset $\mathcal{D}=\{\boldsymbol{X}^{(1)},\cdots,\boldsymbol{X}^{(n)} \}$, where each sample is drawn from $p(\boldsymbol{X})$.


\textbf{Unconditional Generation}: Since the ground truth of $p(\boldsymbol{X})$ is inaccessible, we aim to approximate it with a proxy $q_\theta(\boldsymbol{X})$ parameterized by trainable parameters $\theta$. This allows us to generate new time series samples by drawing from the learned distribution, i.e., $\boldsymbol{X}^{\left ( \text{new} \right )} \sim q_\theta(\boldsymbol{X}) \approx p(\boldsymbol{X})$. 




\textbf{Conditional Generation}: In some cases, we expect that the generated time series are guided by some given condition $\boldsymbol{y} \in \mathbb{R}^C$ (e.g., class labels, observed values). Mathematically, we are now interested in a conditional data distribution $p(\boldsymbol{X}\mid\boldsymbol{y})$. If we can model this conditional distribution with a proxy conditional one $q_\theta(\boldsymbol{X}\mid\boldsymbol{y})$, then we can generate a conditional time series from it as well. 


\subsection{Our Scope}

In this work, we focus on three major and common time series generation tasks. To make it clear, we use the word  \textit{generation} to refer to general time series generation, which encompass synthesis, forecasting, and imputation.


\textbf{Time Series Synthesis}: 
Time series synthesis aims to simulate an entirely new but realistic time series that closely resembles the original data, through learning the underlying real data distribution. It can be subdivided into to 1) unconditional synthesis and 2) class label-guided synthesis. The former one is the same as unconditional generation as previously mentioned, while the latter goals at synthesizing time series that are consistent with the characteristics of a given category, based on the provided class label. Typically, it is implemented via conditional generative models. Given the class label $\boldsymbol{y}\in \mathbb{R}^{C}$, usually represented by categorical encoding methods, e.g., one-hot encoding, our aim is to learn a conditional distribution $q_\theta(\boldsymbol{X}\mid\boldsymbol{y}) \approx p(\boldsymbol{X}\mid\boldsymbol{y})$. This enables controlled synthesis of time series samples belonging to specific categories.




\textbf{Time Series Forecasting}: Time series forecasting targets the prediction of future sequences based on historical observation. Given the observed  time series $\boldsymbol{X}_{\text{obs}} = (\boldsymbol{x}_1, \cdots, \boldsymbol{x}_{L_{\text{obs}}})^\top \in \mathbb{R}^{L_{\text{obs}} \times D}$, we are interested in the conditional distribution $p(\boldsymbol{X}_{\text{pred}} \mid \boldsymbol{X}_{\text{obs}})$, where $\boldsymbol{X}_{\text{pred}} = (\boldsymbol{x}_{L_{\text{obs}}+1}, \cdots, \boldsymbol{x}_{L_{\text{obs}}+L_{\text{pred}}})^\top \in \mathbb{R}^{L_{\text{pred}} \times D}$. After learning the proxy $q_\theta(\boldsymbol{X}_{\text{pred}} \mid \boldsymbol{X}_{\text{obs}})$, we can predict numerous future scenarios by sampling multiple times, $\hat{\boldsymbol{X}}_{\text{pred}} \sim q_\theta(\boldsymbol{X}_{\text{pred}} \mid \boldsymbol{X}_{\text{obs}})$.


\textbf{Time Series Imputation}: 
The objective of the imputation task is to reconstruct missing data values within a time series based on its known observations. We assume that only a partially masked version of the time series  $\tilde{\boldsymbol{X}} = \boldsymbol{X} \odot \mathbf{m}$ is observed, where $\mathbf{m} \in \{0, 1\}^{T \times D}$ is a binary mask matrix indicating missingness. In this context, we are interested in the conditional distribution $p(\boldsymbol{X} \mid \tilde{\boldsymbol{X}}, \mathbf{m})$, so that we can reconstruct the time series by sampling from the fitted proxy $\boldsymbol{X}^\prime \sim q_\theta(\boldsymbol{X} \mid \tilde{\boldsymbol{X}}, \mathbf{m})$, and assemble through $\hat{\boldsymbol{X}} =  {\boldsymbol{X}}  \odot \mathbf{m} + {\boldsymbol{X}^\prime}  \odot (\mathbf{1} -\mathbf{m} )$.


In this paper, we focus on these three widely researched tasks, though there are other generative time series tasks like refinement (super-resolution) \cite{kollovieh2023predict, wang2025a}, which currently account for a tiny proportion. We will also keep on developing \texttt{GenTS} if there are emerging related research topics.



\section{Overview of Generative Time Series Models and Libraries}\label{chap:modeloverview}
\subsection{Taxonomy of Models} 
Time series generative models can be mainly categorized into GANs, VAEs, Diffusions, Flows, and Differential Equations. The former four are typical generative models in general, while the last one is investigated in the time series community and the dynamic system community.

\textbf{GAN-based Models}: Generative Adversarial Networks (GANs) are one of the earliest generative deep learning models \cite {goodfellow2014generative}. It consists of a generator and a discriminator to learn the data distribution directly through adversarial training. To consider the temporal characteristics of time series, some GAN-based variants are implemented through Recurrent Neural Networks (RNN) \cite{esteban2017real} and Attention modules \cite{jeha2022psa}. To further enhance its expressive power, researchers customized adversarial training for time series by introducing additional loss functions \cite{yoon2019time} and integrating continuous time-flow processes \cite{jeon2022gt}.


\textbf{VAE-based Models}: Compared to GANs, Variational Auto Encoder (VAE) has the privilege of stable training and a solid theoretical base \cite{kingma2013auto}, aiming to optimize the Evidence Lower Bound (ELBO). To reconstruct high-quality time series, some VAE-based variants accommodate signal processing and time-frequency modules \cite{desai2021timevae, lee2023vector}. Besides, to consider time series driven by latent dynamics, some variants such as KoVAE \cite{naiman2023generative} combine dynamics theories to reform more informative priors in VAEs.


\textbf{Diffusion-based Models}: As the extension of VAEs \cite{luo2022understanding}, diffusion models for generation tasks have become a major focus of current research. Trained by denoising the corrupted data, diffusion models generate high-quality data samples from pure random noise by means of a gradual denoising process \cite{ho2020denoising}. Similar to VAEs, seasonal decompositions and time-frequency modules are often incorporated to synthesize high-fidelity time series \cite{crabbe2024time, yuan2024diffusion, naiman2024utilizing, wang2025a}. For time series forecasting and imputation, diffusion models have been further improved by modifying the conditional training process \cite{tashiro2021csdi, li2024transformer}.

\textbf{Flow-based Models}: Flow-based models \cite{papamakarios2017masked, dinh2016density, kingma2016improved} map an easily-sampled distribution, such as a standard Gaussian, to a more complex data distribution through a series of invertible transformations. Compared to GANs and VAEs, flow-based models explicitly learn data distributions. In time series generation tasks, they are adapted via modified coupling layers \cite{alaa2021generative} that can satisfy the reversibility and computability of Jacobian determinants.


\textbf{Differential Equation-based Models}: The core concept of differential equation-based time series generative models lies in utilizing ordinary or stochastic differential equations to characterize the intrinsic dynamics of data in continuous time \cite{chen2018neural, patricksde}. These models formalize the evolution of latent states through differential equations parameterized by neural networks, and usually coupled with VAEs and GANs, thereby generating regular or irregular sampled time series data with realistic temporal dependencies and explicit models of temporal dynamics \cite{rubanova2019latent, li2020scalable, kidger2021neural, zhou2023deep}. 

\subsection{Related Benchmark Libraries}


Despite the prosperity of the development of generative time series models, the related benchmark libraries are scarce compared to other fields of generative AI. In the time series community, the existing benchmark libraries rarely provide comprehensive inclusion of multi-domain datasets, various SOTA generative models, different generation tasks, and panoramic evaluation metrics at the same time, as shown in Table.~\ref{tab:restructured_comparison}. 

\textbf{Coverage of Generative Time Series Models and Tasks}: As we investigated, most of the pioneering time series libraries either oriented towards discriminative models, or ignored the versatility of generative time series models. For example, TSLib and GluonTS provides a collection of discriminative models on classical time series analysis tasks, but lacks any implementations of generative models. Meanwhile, including typical generative time series models, TSGM \cite{nikitin2024tsgm} and Synthcity \cite{qian2023synthcity} only support unconditional synthesis while ProbTS merely focuses on time series forecasting. It should be also mentioned that two libraries, TSGBench \cite{nikitin2024tsgm} and CTBench \cite{qian2023synthcity}, merely address the evaluation of generated time series, respectively from domain-agnostic and domain-specific perspectives, neglecting the data processing and model training. Therefore, the existing related libraries fail to cover the broader spectrum of time series generation that are critical for real-world applications.

\textbf{Extensibility of the Existing Libraries for Generative Models}: In addition, it is objectively difficult to incorporate generative time series models into the existing libraries in terms of implementation. On the one hand, since traditional time series libraries, like TSLib and GluonTS, focus on discriminative models instead of generative ones, their training objectives could be straightforward, e.g., MSE/quantile loss for regression and cross entropy for classification. The accessible application programming interfaces (APIs) for user customization are thus also designed for discriminative models. However, generative time series models could have complex and various training paradigms as we introduced above, such as adversarial training, diffusion-denoising, making it difficult to fit generative models in those libraries with APIs for discriminative models. On the other hand, existing generative time series libraries, as mentioned previously, primarily stick to either unconditional synthesis or time series forecasting. They also lack standardized data processing pipelines for various tasks, as well as the APIs to incorporate different types of  exogenous conditions, e.g. categorical class labels or temporal observation masks. 

Therefore, we are motivated to proposed a comprehensive and customizable benchmark library for the community of generative time series models, which has unified data preparation, flexible model training, and thorough performance evaluation.

\begin{figure*}[htb]
  \centering
  \includegraphics[width=\linewidth]{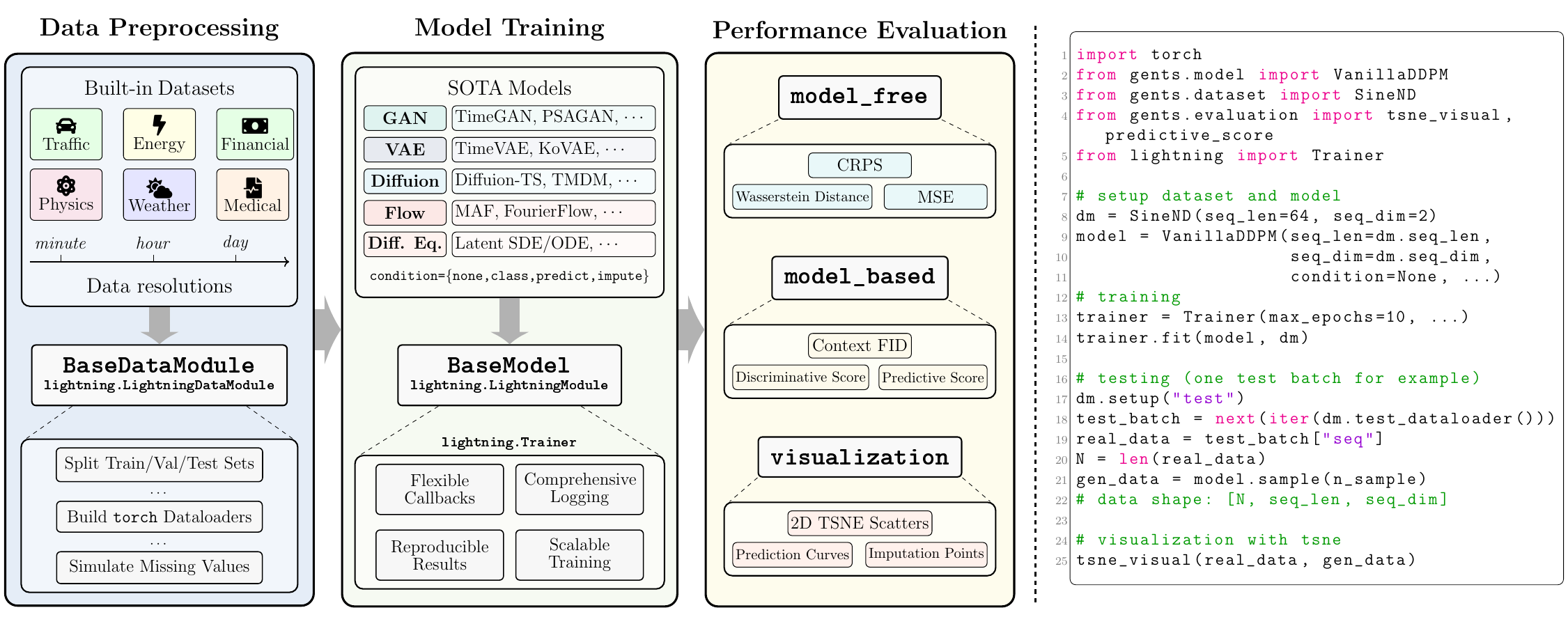}
  \caption{\textit{Left}: The overview of \texttt{GenTS} framework. \textit{Right}: A code snippet showing the neat usage of \texttt{GenTS} in less than 30 lines.}
  \label{fig:framework}
  \Description{A framework about GenTS including Data Preprocessing, Model Training, and Performance Evaluation. A code snippet showing how to use GenTS library.}
\end{figure*}

\section{\texttt{GenTS}}

Our proposed comprehensive library, \texttt{GenTS}, will be introduced in this section, with a modular design covering the full benchmarking pipeline, i.e., data preprocessing, model training, and performance evaluation. Built on \texttt{torch} and \texttt{lightning}\footnote{Also known as \texttt{pytorch-lightning}}, \texttt{GenTS} is highly user-friendly for researchers to benchmark the existing models and datasets, and facilitate the  realization of new ideas. Figure~\ref{fig:framework} illustrates the structure of \texttt{GenTS} alongside a corresponding code snippet for a quick start. 


\subsection{Data Preprocessing}

\textbf{Multi-domain Datasets} Our \texttt{GenTS} library features a collection of over fifteen time series datasets from six different domains, including traffic, energy, financial, and others. Most of them can be publicly accessed by web downloading and recorded in \texttt{.csv} / \texttt{.txt} / \texttt{.tsf} file formats. Our \texttt{BaseDataModule} parses all these files into tensors of time series windows and builds corresponding dataloaders for training, validation, and testing.

In addition to the widely used real-world data, we simulated two simple physical trajectory data (\texttt{Spiral2D} and \texttt{SineND}, see details in \ref{apendix:data}) for users to quickly test their ideas of new models. As shown in Table.~\ref{tab:datasets}, these time series datasets range from less than 1 minute to 1 day, presenting distinct temporal patterns. Therefore, the multi-domain datasets in our library exhibit strong diversity that enables researchers to benchmark and develop generative time series models for various applications.

\begin{table}[tb]
\centering
\caption{Built-in Time Series Datasets in GenTS}
\label{tab:datasets}
\begin{tabularx}{\linewidth}{lcccc}
\toprule
\textbf{Name} & \textbf{Resolution} & \textbf{No. of Variables} & \textbf{Domain} \\
\midrule
SineND & continuous & $N$ (user-configurable) & Physics \\
Spiral2D & continuous & 2 & Physics \\
Stocks & 1 day & 6 & Financial \\
Energy & 10 min & 28 & Energy \\
ETT & 1 hour/15 min & 7 & Energy \\
Electricity & 1 hour & 321 & Energy \\
Traffic & 1 hour & 862 & Traffic \\
Exchange & 1 day & 8 & Financial \\
MoJoCo & continuous & 14 & Physics \\
Physionet & 1 min - 1 hour & 35 & Medical \\
ECG & $\sim$700 Hz & 1 & Medical \\
Air quality & 1 hour & 6 & Weather \\
Weather & 10 min & 6 & Weather \\
\bottomrule
\end{tabularx}
\end{table}

\textbf{Customizable Base Data Module} To realize data preparation for different time series generation tasks, we designed a base class named \texttt{BaseDataModule}. Inherited from \texttt{LightningDataModule}, it not only performs fundamental functions like dataset splitting and dataloader instantiation, but also completes task-related jobs, for example, simulating missing values for time series imputation, or assigning look-back windows for time series forecasting. Besides, for some models dealing with irregular time series, e.g., neural differential equations, we can optionally simulate irregular dropping of partial time steps. More importantly, users can effortlessly construct their own time series datamodule beyond our built-in datasets. By following the structure of \texttt{BaseDataModule}, users can implement their own specialized data preprocessing pipelines. The detailed introduction on how to customize \texttt{BaseDataModule} can be found in the tutorial on our document website.




\begin{table}[tb]
    \centering
    \caption{GenTS Models. Syn.: Synthesis ("+" indicates  supporting class labels), Fcst: Forecasting, Impt.: Imputation}
    \label{tab:models}
\begin{tabularx}{\linewidth}{lc *{3}{>{\centering\arraybackslash}X}}
\toprule
\textbf{Model} & \textbf{Type} & \textbf{Syn.} & \textbf{Fcst.} & \textbf{Impt.}  \\
\midrule
VanillaVAE$^{\ast }$  & \multirow{4}{*}{\centering VAE} 
  & \checkmark + & \checkmark & \checkmark  \\
TimeVAE \cite{desai2021timevae}           
  &                                           
  & \checkmark & ---        & ---               \\
TimeVQVAE \cite{lee2023vector}             
  &                                           
  & \checkmark + & ---        & ---         \\
KoVAE \cite{naiman2023generative}          
  &                                           
  & \checkmark & ---        & ---                \\
\midrule
VanillaGAN$^{\ast }$  & \multirow{6}{*}{\centering GAN}
  & \checkmark + & \checkmark & \checkmark  \\
COSCIGAN \cite{seyfi2022generating}        
  &                                           
  & \checkmark & ---        & ---               \\
TimeGAN \cite{yoon2019time}                 
  &                                           
  & \checkmark & ---        & ---                \\
GTGAN \cite{jeon2022gt}                    
  &                                           
  & \checkmark & ---        & ---                \\
PSAGAN \cite{jeha2022psa}                  
  &                                           
  & \checkmark & ---        & ---                \\
RCGAN \cite{esteban2017real}                
  &                                           
  & \checkmark + & ---        & ---         \\
\midrule
VanillaMAF$^{\ast }$  & \multirow{2}{*}{\centering Flow}
  & \checkmark + & \checkmark & \checkmark  \\
FourierFlow$^{\#}$ \cite{alaa2021generative}                                   
  &                                           
  & \checkmark & ---        & ---              \\
\midrule
VanillaDDPM$^{\ast }$  & \multirow{7}{*}{\centering Diffusion}
  & \checkmark + & \checkmark & \checkmark  \\
CSDI \cite{tashiro2021csdi}                  
  &                                           
  & ---        & \checkmark & \checkmark        \\
DiffusionTS \cite{yuan2024diffusion}        
  &                                           
  & \checkmark & \checkmark & \checkmark         \\
TMDM \cite{li2024transformer}                
  &                                           
  & ---        & \checkmark & ---                \\
FourierDiffusion \cite{crabbe2024time}       
  &                                           
  & \checkmark & ---        & ---                \\
ImagenTime \cite{naiman2024utilizing}        
  &                                           
  & \checkmark & \checkmark & \checkmark         \\
FIDE$^{\#}$ \cite{galib2024fide}                    
  &                                           
  & \checkmark & ---        & ---                \\
\midrule
LatentODE \cite{rubanova2019latent} & \multirow{4}{*}{\centering Diff. Eq.}
  & \checkmark & \checkmark & \checkmark  \\
LatentSDE \cite{li2020scalable}     
  &                                           
  & \checkmark & ---        & ---            \\
SDEGAN \cite{kidger2021neural}       
  &                                           
  & \checkmark & ---        & ---            \\
LS4 \cite{zhou2023deep}              
  &                                           
  & \checkmark & \checkmark & \checkmark  \\
\bottomrule
\multicolumn{5}{p{\linewidth}}{\footnotesize $^{\ast }$ denotes simple baselines implemented by the authors; $^{\# }$ denotes models that are only applicable to univariate time series.} 
\end{tabularx}
\end{table}

\subsection{Model Training}

\textbf{Built-in Models} As mentioned in the previous overview in Section~\ref{chap:modeloverview}, we covered five main generative model types in \texttt{GenTS}, currently including more than 25 SOTA models (and their variants) from top conferences. Our implementation of these models mostly sticks to their respective official codes, except for some equivalent adaptations, e.g., from \texttt{tensorflow} to \texttt{torch}.  In addition to these SOTA models, we implemented four naive baselines, i.e., \texttt{VanillaVAE}, \texttt{VanillaGAN}, \texttt{VanillaDDPM}, and \texttt{VanillaMAF}, which are based on either simple MLP backbones or the 1-D version of computer vision backbones. They are the direct usage of generative models without any adaptation to time series data, offering a reasonable starting point for developing new ideas. All of our existing models and their respective capabilities are listed in Table.~\ref{tab:models}, and the more details of our Vanilla models can be found in the Appendix.~\ref{chap:model}.

\textbf{Customizable Base Model Module} Similar to \texttt{BaseDataModule}, a base model class \texttt{BaseModel} is designed as a template for all models. In this framework, we inherit \texttt{LightningModule} and then respectively implement \texttt{training\_step} and \texttt{validation\_step} for each model. Since some GAN-based models, like TimeGAN, lack validation during model training in their official codes, we supplement the Wasserstein distance as the validation metric for them as well. Utilizing \texttt{lightning.Trainer}, users can effortlessly fulfill the training loop and efficiently realize flexible callbacks, comprehensive logging, gradient accumulation, and other useful functions during the model training process. Meanwhile, we design a \texttt{sample} method in \texttt{BaseModel} to generate time series after fitting. Given the number of sampling times and the optional conditions, the \texttt{sample} method returns the batched generated time series. Most importantly, owing to the modular model design, researchers can also customize their own generative time series model based on the \texttt{BaseModel}, as long as they implement the necessary training/validation/sampling process as well. We detailed how to customize generative time series models on our document website.



\subsection{Performance Evaluation}
Unlike classical time series analysis tasks, which usually have clear evaluation metrics, e.g., forecasting accuracy, the evaluation of time series generation can be multi-dimensional, depending on the specific contexts. In \texttt{GenTS}, we provide both model-free and model-based metrics to comprehensively evaluate generated time series data for different goals.

\textbf{Model-free Metrics} Without training additional neural networks, model-free metrics evaluate the quality of generated time series from a statistical perspective. For example, we include the Wasserstein distance (and its variants) \cite{bonneel2015sliced}, the Continuous Ranked Probability Score (CRPS), Mean Squared Error (MSE), and others. These metrics comprehensively measure the distributional closeness between the generated time series and the real ones.

\textbf{Model-based Metrics} Leveraging auxiliary neural networks, model-based metrics evaluate the fidelity and usefulness of generated time series through the lens of neural network perception, usually in the context of time series synthesis task. Specifically, we include Predictive Score (PS), Discriminative Score (DS) \cite{yoon2019time}, and Context-FID (C-FID) \cite{jeha2022psa} as metrics. The former one evaluates the usefulness of generated time series as a training dataset for the time series forecasting task, while the latter two focus on the fidelity based on how indistinguishable or latently close the generated data are to the real one. In this way, the quality of the synthetic time series is evaluated based on how neural networks learn and represent temporal dynamics. More details can be found in Appendix.~\ref{chap:metrics}.

\textbf{Visualization}
Beyond quantitative evaluation, we provide qualitative visualization in \texttt{GenTS} to compare the generated time series with the real ones. Specifically, t-SNE is included for visualizing time series in a 2D space, commonly used to identify whether the learned data distribution is close to the real one. For time series forecasting and imputation, we provide useful functions to plot forecasting and imputation results in one call, where both the averaged predictions and the prediction intervals will be demonstrated.




\section{Experiments}
In this section, we benchmarked the generative time series models (Table.~\ref{tab:models}) on multi-domain datasets (Table.~\ref{tab:datasets}) via our \texttt{GenTS} over three time series generation tasks, i.e., synthesis (with/without class label guidance), forecasting, and imputation. 

\subsection{Time series synthesis} \label{chap:main_exp}
\begin{figure*}[htb]
    \centering
    \resizebox{\textwidth}{!}{\includegraphics[height=1.cm]{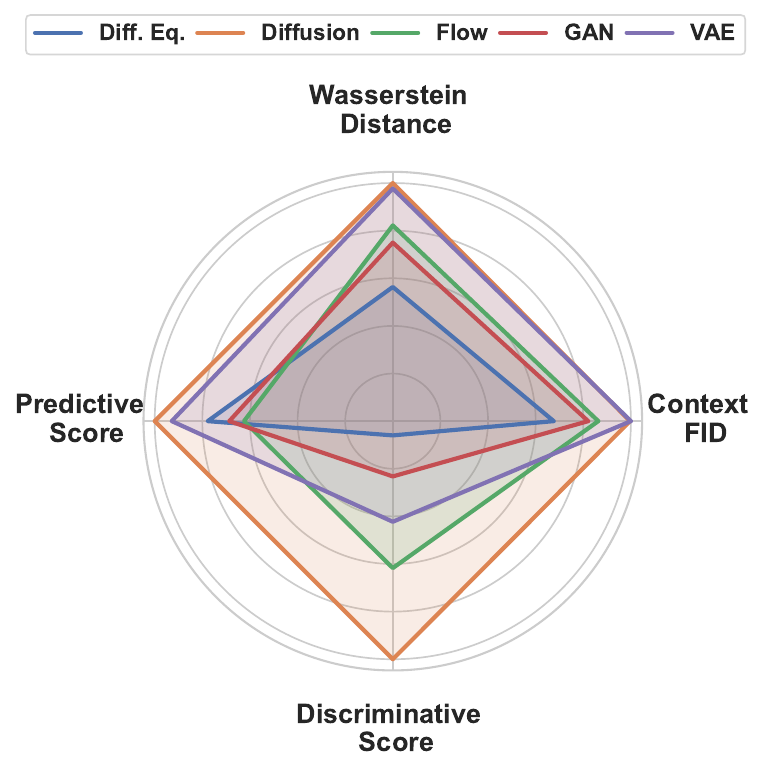}
    \includegraphics[height=1.cm]{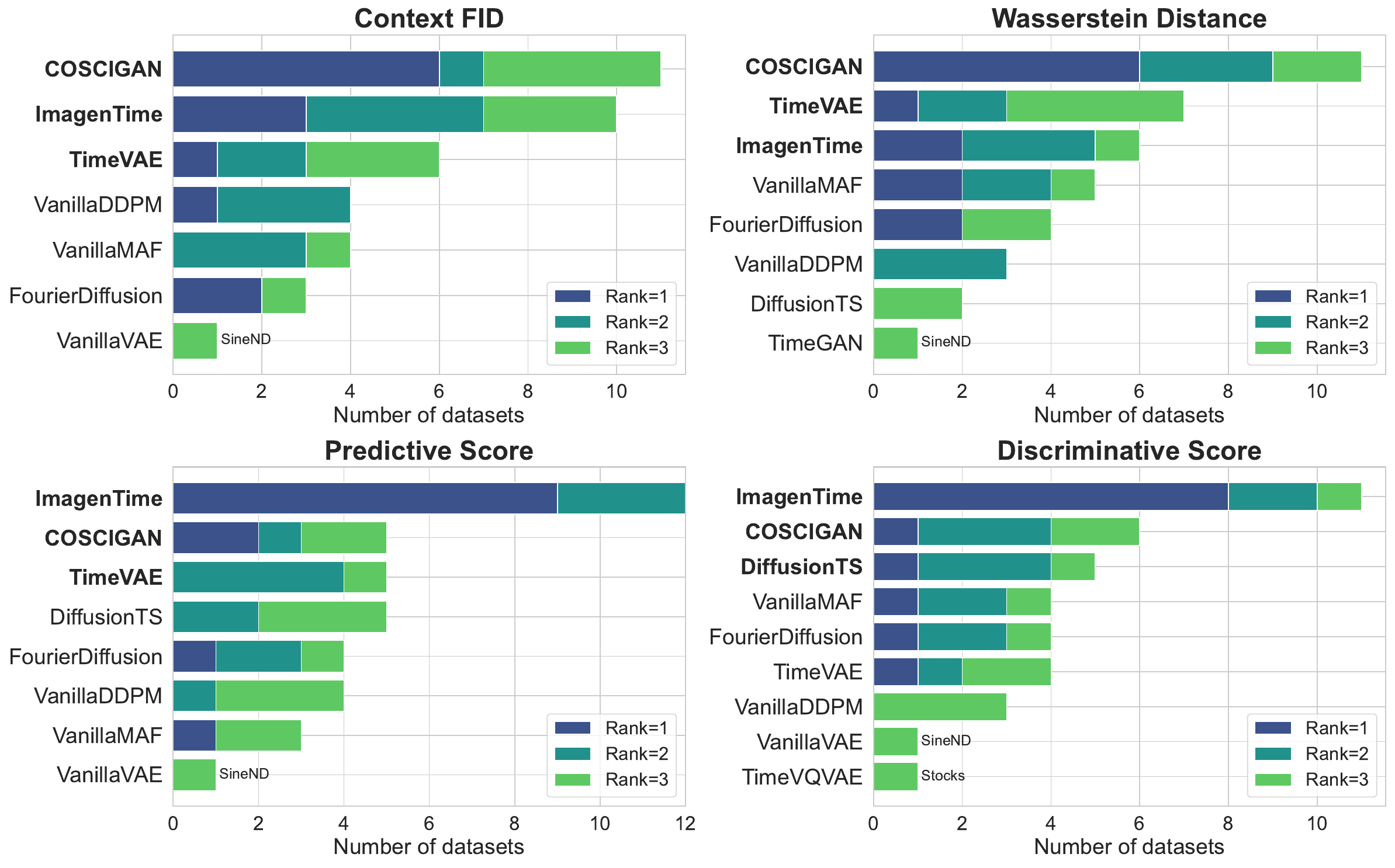}}
    
    \caption{Main Results of Time Series Synthesis Benchmarking. \textit{Left}: Averaged normalized performance of each type of model over four metrics. The greater, the better. \textit{Right}: Top-3 counts of each model over all datasets.}
    \Description{The radar chart records overall performance of different types of model families, where Diffusions are the best. The bar charts reveal the number of top-3 counts of specific models across all datasets.}
    \label{fig:syn_exp}
\end{figure*}

\textbf{Setting} In this task, we aim to test the benchmarks' capability of fitting the complex data distributions. Specifically, for each dataset, we fixed the length of the time series as $T=24$, in the multivariate setting $D=\min(16, D_N)$, where $D_N$ denotes the total number of channels in the dataset. Since some models are designed for univariate time series, we also launched univariate synthesis experiments (attached in the appendix \ref{chap:exp_appendix}). For class label-guided generation, we selected the datasets and models that are compatible with class labels. All the models are evaluated by Wasserstein Distance, C-FID, PS, and DS. Please see our appendix \ref{chap:exp_appendix} for more experiment details.

\textbf{Results of Unconditional Synthesis} As Figure~\ref{fig:syn_exp} depicts (with full results in appendix \ref{chap:univar_exp}), different types of models generally exhibited varying performances on the evaluation metrics. Diffusion models showed overall superiority, especially on the model-based metrics, i.e., PS and DS. VAEs also demonstrated competitive performances on Wasserstein Distance and C-FID, implying their capability of modeling realistic high-dimensional distributions. 

In terms of individual model performance, ImagenTime, COSCIGAN, and TimeVAE showed advantages. Specifically, ImagenTime aligned with the overall performance of diffusion-based models, excelling at PS and DS. Despite the general inferior performance of GAN-based models, COSCIGAN stood out and showed superiority on C-FID and Wasserstein Distance. TimeVAE achieved balanced but moderate performances in general. The univariate results can be found in the appendix \ref{chap:univar_exp}. 

\begin{figure*}[htb]
    \centering
    \includegraphics[width=\linewidth]{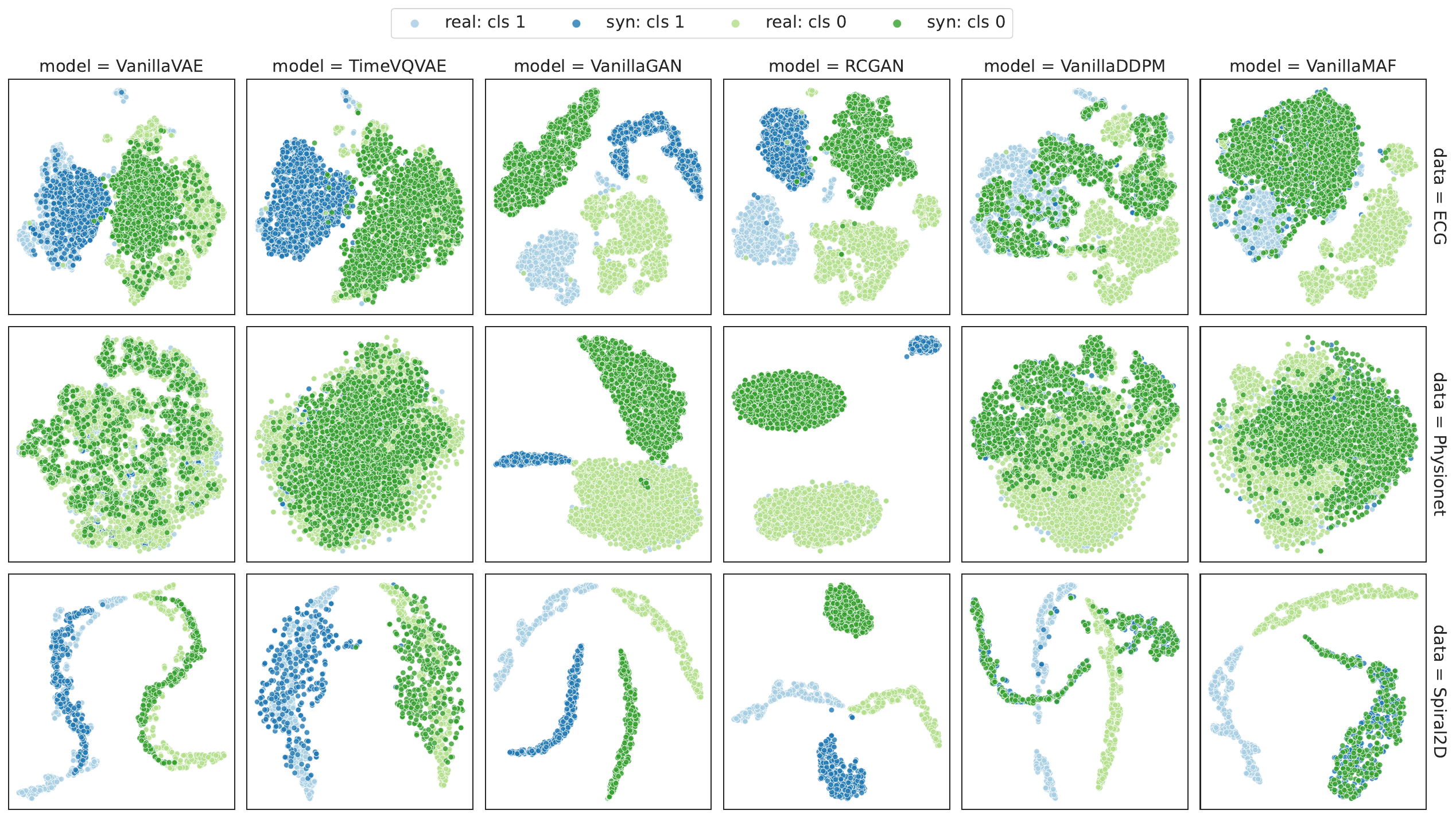}
    \caption{t-SNE Visualization of Class Label-guided Time Series Generation. In ECG dataset, we selected first two classes from the total five classes in the ECG dataset for visualization purposes.}
    \label{fig:syn_cls}
    \Description{The scatter plots shows the how well different models fit on the data distribution. Different colors represent different labels of time series data.}
\end{figure*}

\textbf{Results of Class Label-Guided Synthesis}. Table.~\ref{tab:syn_cls} and Figure~\ref{fig:syn_cls} respectively demonstrate the quantitative and qualitative benchmarking results. TimeVQVAE's superior performance was highlighted in terms of statistical and representative closeness to the labeled time series data, showing the most aligned clusters with the real class-labeled dataset. In contrast, though GANs generated clearly categorized time series, they are not well aligned with the real clusters. In terms of Physionet, although VanillaMAF and VanillaDDPM  demonstrated slight advantages, we reckoned that the six models may not fit the real data distribution well, as their DS all approached 0.5. This indicated that the generated data can be easily distinguished from the real ones by post-hoc trained classifiers.


\begin{table}[htb]
    \centering
    \caption{Full Results of Class Label-Guided Time Series Synthesis.}
    \label{tab:syn_cls}
    
\begin{tabular}{@{}l|c|cccc@{}}
\toprule
\multirow{2}{*}{\textbf{Dataset}}      & \multirow{2}{*}{\textbf{Model}} & \multicolumn{4}{c}{\textbf{Metrics}}    \\
                           &                        & C. FID & W. D. & P. S. & D. S. \\ \midrule
\multirow{6}{*}{ECG}       & VanillaVAE             & 0.557  & 0.289 & \textbf{0.189} & 0.442 \\
                           & TimeVQVAE              & \textbf{0.069}  & \textbf{0.101} & 0.191 & \textbf{0.311} \\
                           & VanillaGAN             & 0.867  & 0.492 & 0.672 & 0.496 \\
                           & RCGAN                  & 1.160  & 0.425 & 0.295 & 0.486 \\
                           & VanillaDDPM            & 0.202  & 0.256 & 0.238 & 0.457 \\
                           & VanillaMAF             & 0.799  & 0.613 & 0.373 & 0.491 \\ \midrule
\multirow{6}{*}{Physionet} & VanillaVAE             & 10.744 & 0.306 & 0.166 & 0.499 \\
                           & TimeVQVAE              & 8.015  & 0.286 & 0.172 & 0.499 \\
                           & VanillaGAN             & 8.280  & 0.439 & 0.200 & 0.500 \\
                           & RCGAN                  & 8.081  & 0.314 & 0.179 & 0.500 \\
                           & VanillaDDPM            & \textbf{3.310}  & \textbf{0.153} & 0.149 & 0.497 \\
                           & VanillaMAF             & 11.370 & 0.353 & \textbf{0.146} & \textbf{0.496} \\ \midrule
\multirow{6}{*}{Spiral2D}  & VanillaVAE             & 1.300  & 0.224 & 0.412 & 0.282 \\
                           & TimeVQVAE              & \textbf{0.974}  & \textbf{0.139} & \textbf{0.408} & 0.239 \\
                           & VanillaGAN             & 4.531  & 0.449 & 0.417 & 0.404 \\
                           & RCGAN                  & 4.964  & 0.556 & 0.432 & 0.282 \\
                           & VanillaDDPM            & 1.721  & 0.286 & 0.413 & \textbf{0.075} \\
                           & VanillaMAF             & 8.469  & 0.793 & 0.516 & 0.464 \\ \bottomrule
\end{tabular}
\end{table}


\subsection{Time series forecasting}
\textbf{Setting} We followed one of the most widely used time series forecasting setting \cite{wang2024deep}, and set both look-back window and prediction window length as 96, i.e. $L_{\text{obs}} = L_{\text{pred}}=96$, in the multivariate context. Each trained model was inferred 50 times to assemble probabilistic forecasts, and the deterministic ones were derived by averaging over the sampled 50 predictions. The forecasts were evaluated by MSE and CRPS, corresponding to point and distributional accuracy.

\textbf{Results} Table.~\ref{tab:fcst} provides a complete evaluation via metrics of MSE and CRPS. The results revealed that CSDI and TMDM frequently rank among the best for both deterministic and probabilistic forecasting. It should be noted that in Stocks, a time series dataset with non-stationarity, TMDM showed absolute advantages over the others, thanks to its backbone \cite{liu2022non}, specializing in handling non-stationarity in time series forecasting.

We noticed that two baselines, VanillaMAF and VanillaVAE, also demonstrated competitive performance across some datasets, especially on the ETTh, Electricity, and Spiral2D. This implied that for some cases of time series forecasting, less architectural complexity may provide satisfying baselines as well. It may guide users to explore their potential in specific applications in vertical industries.

\begin{table*}[htb]
    \centering
     \caption{Full results of time series forecasting models on MSE and CRPS metrics. The best model on each dataset is in bold; the second and third-best are underlined. The last row reports the frequency of a model ranking top3.}
    \label{tab:fcst}
    \resizebox{\textwidth}{!}{
    \begin{tabular}{l | cc | cc | cc | cc | cc | cc | cc | cc | cc | cc | 
    cc}
\toprule
Dataset & \multicolumn{2}{c}{\textbf{CSDI}} & \multicolumn{2}{c}{\textbf{DiffusionTS}} & \multicolumn{2}{c}{\textbf{ImagenTime}} & \multicolumn{2}{c}{\textbf{LS4}} & \multicolumn{2}{c}{\textbf{LatentODE}} & \multicolumn{2}{c}{\textbf{PSAGAN}} & \multicolumn{2}{c}{\textbf{TMDM}} & \multicolumn{2}{c}{\textbf{VanillaDDPM}} & \multicolumn{2}{c}{\textbf{VanillaGAN}} & \multicolumn{2}{c}{\textbf{VanillaMAF}} & \multicolumn{2}{c}{\textbf{VanillaVAE}} \\
  & MSE & CRPS & MSE & CRPS & MSE & CRPS & MSE & CRPS & MSE & CRPS & MSE & CRPS & MSE & CRPS & MSE & CRPS & MSE & CRPS & MSE & CRPS & MSE & CRPS \\
\midrule
AirQuality & \uline{0.436} & \uline{0.153} & 0.664 & 0.200 & \textbf{0.322} & \textbf{0.137} & 0.724 & 0.269 & 0.599 & 0.242 & 0.689 & 0.213 & 0.759 & 0.224 & \uline{0.557} & \uline{0.188} & 0.742 & 0.272 & 0.690 & 0.219 & 0.582 & 0.196 \\
ECG & 0.181 & \uline{0.097} & 0.205 & 0.111 & 0.245 & 0.104 & 2.259 & 0.520 & 0.295 & 0.151 & 0.712 & 0.285 & 0.193 & 0.110 & \textbf{0.149} & \uline{0.094} & 1.067 & 0.382 & \uline{0.173} & \textbf{0.091} & \uline{0.179} & 0.099 \\
ETTh1 & 0.729 & 0.260 & \uline{0.611} & \uline{0.246} & 0.625 & 0.250 & 3.671 & 0.727 & 0.963 & 0.360 & 1.652 & 0.428 & 0.844 & 0.266 & 0.714 & 0.272 & 1.790 & 0.524 & \textbf{0.541} & \textbf{0.218} & \uline{0.556} & \uline{0.239} \\
ETTh2 & 0.190 & \uline{0.119} & 0.173 & 0.126 & \uline{0.154} & 0.120 & 2.748 & 0.641 & 0.343 & 0.211 & 0.406 & 0.206 & 0.143 & \textbf{0.111} & 0.794 & 0.309 & 1.817 & 0.532 & \uline{0.148} & 0.117 & \textbf{0.135} & \uline{0.120} \\
Electricity & \uline{0.229} & \uline{0.121} & 0.265 & 0.149 & 0.277 & 0.149 & 2.652 & 0.530 & 0.397 & 0.211 & 1.076 & 0.323 & \textbf{0.198} & \textbf{0.119} & 0.358 & 0.175 & 1.193 & 0.386 & \uline{0.256} & \uline{0.138} & 0.264 & 0.159 \\
Energy & \textbf{0.427} & \textbf{0.188} & \uline{0.580} & \uline{0.259} & 0.831 & 0.323 & 1.697 & 0.514 & 0.973 & 0.378 & 2.221 & 0.558 & \uline{0.533} & \uline{0.204} & 0.869 & 0.344 & 2.317 & 0.589 & 0.891 & 0.342 & 0.759 & 0.321 \\
Exchange & \textbf{0.078} & \textbf{0.076} & 0.790 & 0.357 & 0.441 & \uline{0.235} & 2.904 & 0.671 & 0.741 & 0.340 & 1.004 & 0.393 & \uline{0.117} & \uline{0.111} & 0.658 & 0.345 & 3.380 & 0.617 & \uline{0.394} & 0.243 & 0.609 & 0.319 \\
MuJoCo & \textbf{2.024} & \textbf{0.264} & 2.448 & 0.342 & 2.784 & 0.385 & 76.755 & 2.195 & 2.217 & 0.408 & 3.725 & 0.530 & 2.255 & \uline{0.320} & 3.105 & 0.429 & 4.341 & 0.656 & \uline{2.130} & \uline{0.296} & \uline{2.135} & 0.323 \\
SineND & 0.128 & \uline{0.114} & 0.202 & 0.152 & \uline{0.124} & \uline{0.115} & 0.771 & 0.354 & 0.125 & 0.151 & 0.132 & 0.159 & \textbf{0.032} & \textbf{0.056} & 0.133 & 0.118 & 0.173 & 0.171 & \uline{0.124} & 0.116 & 0.125 & 0.153 \\
Spiral2D & 0.005 & 0.019 & 0.040 & 0.069 & 0.008 & 0.026 & 2.519 & 0.629 & 0.016 & 0.037 & 1.099 & 0.415 & 0.001 & 0.016 & \textbf{0.001} & \uline{0.010} & 0.688 & 0.319 & \uline{0.001} & \textbf{0.009} & \uline{0.001} & \uline{0.010} \\
Stocks & 2.775 & 0.732 & 6.976 & 1.240 & 4.724 & 0.970 & 3.003 & 0.827 & 4.436 & 0.947 & \uline{0.697} & \uline{0.337} & \textbf{0.156} & \textbf{0.143} & 2.462 & 0.749 & 9.323 & 1.448 & 4.733 & 0.979 & \uline{0.734} & \uline{0.381} \\
Traffic & \textbf{0.367} & \textbf{0.113} & 0.748 & 0.185 & \uline{0.375} & \uline{0.119} & 2.268 & 0.503 & 0.442 & 0.158 & 2.111 & 0.465 & 1.603 & 0.381 & 0.434 & 0.134 & 1.372 & 0.385 & 0.416 & \uline{0.122} & \uline{0.379} & 0.139 \\
Weather & \uline{0.239} & \uline{0.103} & 0.248 & 0.114 & 0.283 & 0.122 & 82.509 & 0.890 & 0.371 & 0.186 & 0.527 & 0.185 & \textbf{0.228} & \textbf{0.102} & 2.556 & 0.491 & 1.805 & 0.517 & 0.789 & 0.121 & \uline{0.231} & \uline{0.112} \\
\midrule
\textbf{\textit{Top3 Count}} & 7 & \textbf{10} & 2 & 2 & 4 & 4 & 0 & 0 & 0 & 0 & 1 & 1 & 6 & 8 & 3 & 3 & 0 & 0 & \textbf{8} & 6 & \textbf{8} & 5 \\
\bottomrule
\end{tabular}
}
   
\end{table*}

\subsection{Time series imputation}
\textbf{Setting} In the task of imputation, we set the length of time series as $L=24$, and simulated missing data at the missing rate of $0.2$. All the models were also inferred 50 times to have the probabilistic and deterministic predictions the same as the time series forecasting setting.

\textbf{Results} Table.~\ref{tab:imp} records the full results of time series imputation. Different from previous tasks, Diffusion-based models dominated in this task with absolute superiority over all the datasets. Specifically, CSDI consistently outperformed on most of the tested datasets, followed by ImagenTime and DiffusionTS. This indicated that diffusion-based generative models are particularly effective for time series imputation, offering a strong balance of both accurate point predictions and well-calibrated uncertainty estimates. In contrast, Differential Equation-based models and naive baselines generally exhibited weaker performance, rarely achieving top-3 rankings and often showing higher imputation errors.

\begin{table*}[htb]
\centering
\caption{Full results of time series imputation models on MSE and CRPS Metrics. The best model on each dataset is in bold; the second and third-best are underlined. The last row reports the frequency of a model ranking top3.}
\label{tab:imp}
\resizebox{\textwidth}{!}{
\begin{tabular}{l | cc | cc | cc | cc | cc | cc | cc | cc | cc}
\toprule
Dataset & \multicolumn{2}{c}{\textbf{CSDI}} & \multicolumn{2}{c}{\textbf{DiffusionTS}} & \multicolumn{2}{c}{\textbf{ImagenTime}} & \multicolumn{2}{c}{\textbf{LS4}} & \multicolumn{2}{c}{\textbf{LatentODE}} & \multicolumn{2}{c}{\textbf{VanillaDDPM}} & \multicolumn{2}{c}{\textbf{VanillaGAN}} & \multicolumn{2}{c}{\textbf{VanillaMAF}} & \multicolumn{2}{c}{\textbf{VanillaVAE}} \\
& MSE & CRPS & MSE & CRPS & MSE & CRPS & MSE & CRPS & MSE & CRPS & MSE & CRPS & MSE & CRPS & MSE & CRPS & MSE & CRPS \\
\midrule
AirQuality & \uline{0.734} & 0.243 & 0.737 & 0.236 & 0.902 & 0.209 & \textbf{0.734} & 0.243 & \uline{0.734} & 0.243 & 0.757 & \textbf{0.190} & 0.976 & 0.215 & 0.812 & \uline{0.195} & 0.739 & \uline{0.193} \\
ECG & 0.037 & 0.018 & 0.025 & \uline{0.016} & \textbf{0.018} & \textbf{0.012} & 0.309 & 0.077 & 0.128 & 0.040 & \uline{0.022} & \uline{0.018} & 0.168 & 0.068 & \uline{0.081} & 0.027 & \uline{0.034} & 0.019 \\
ETTh1 & \textbf{0.009} & \textbf{0.010} & \uline{0.017} & \uline{0.015} & \uline{0.015} & \uline{0.013} & 0.390 & 0.091 & 0.072 & 0.040 & 0.052 & 0.033 & 0.296 & 0.087 & 0.051 & 0.029 & 0.040 & 0.027 \\
ETTh2 & \textbf{0.001} & \textbf{0.004} & \uline{0.003} & \uline{0.007} & \uline{0.002} & \uline{0.005} & 0.832 & 0.136 & 0.020 & 0.022 & 0.021 & 0.020 & 0.220 & 0.083 & 0.011 & 0.014 & 0.012 & 0.016 \\
Electricity & \textbf{0.011} & \textbf{0.012} & 0.019 & 0.018 & \uline{0.015} & \uline{0.015} & \uline{0.753} & \uline{0.130} & 0.058 & 0.036 & 0.052 & 0.032 & 0.403 & 0.104 & 0.030 & 0.021 & 0.030 & 0.024 \\
Energy & \textbf{0.006} & \textbf{0.003} & \uline{0.016} & \uline{0.015} & \uline{0.012} & \uline{0.009} & 0.222 & 0.078 & 0.074 & 0.043 & 0.053 & 0.037 & 0.688 & 0.143 & 0.062 & 0.035 & 0.041 & 0.029 \\
Exchange & \textbf{0.000} & \textbf{0.002} & 0.057 & \uline{0.049} & \uline{0.012} & \uline{0.018} & 0.379 & 0.089 & 0.125 & 0.066 & 0.085 & 0.058 & 0.223 & 0.083 & \uline{0.039} & 0.033 & 0.140 & 0.068 \\
MuJoCo & \uline{0.083} & \textbf{0.013} & \textbf{0.049} & \uline{0.018} & 0.292 & \uline{0.056} & 37.556 & 0.640 & 1.254 & 0.126 & 2.581 & 0.180 & 14.275 & 0.434 & \uline{0.627} & 0.071 & 1.700 & 0.140 \\
SineND & \textbf{0.000} & \textbf{0.001} & \uline{0.002} & \uline{0.006} & \uline{0.000} & \uline{0.002} & 0.098 & 0.047 & 0.017 & 0.022 & 0.003 & 0.009 & 0.021 & 0.026 & 0.004 & 0.008 & 0.010 & 0.014 \\
Spiral2D & \uline{0.000} & \uline{0.002} & 0.001 & 0.005 & \textbf{0.000} & \textbf{0.001} & 0.916 & 0.171 & 0.072 & 0.040 & 0.002 & 0.007 & 0.109 & 0.048 & 0.002 & 0.005 & \uline{0.001} & \uline{0.004} \\
Stocks & \uline{0.071} & \uline{0.055} & \textbf{0.022} & \textbf{0.029} & 0.286 & 0.100 & 1.423 & 0.249 & 0.838 & 0.184 & 0.234 & 0.104 & 4.409 & 0.429 & 1.089 & 0.211 & \uline{0.064} & \uline{0.046} \\
Traffic & \textbf{0.026} & \textbf{0.011} & \uline{0.037} & \uline{0.017} & \uline{0.034} & \uline{0.014} & 0.501 & 0.095 & 0.068 & 0.028 & 0.070 & 0.024 & 0.411 & 0.104 & 0.057 & 0.020 & 0.050 & 0.024 \\
Weather & \textbf{0.006} & \textbf{0.003} & \uline{0.008} & \uline{0.004} & \uline{0.009} & \uline{0.004} & 15.008 & 0.144 & 0.024 & 0.019 & 0.051 & 0.027 & 0.955 & 0.144 & 0.020 & 0.010 & 0.098 & 0.051 \\
\midrule
\textbf{\textit{Top3 Count}} & \textbf{12} & \textbf{11} & 9 & \textbf{11} & 10 & \textbf{11} & 1 & 0 & 1 & 0 & 0 & 2 & 0 & 0 & 3 & 1 & 3 & 3 \\
\bottomrule
\end{tabular}}
\end{table*}


\section{Discussions}
In this section, we will offer some suggestions for generative time series models, according to our previous results, and discuss the potential future research directions.

\textbf{Suggestions on model selection}. In terms of all the mentioned time series generation tasks, we can see that \textit{diffusion-based models} were frequently highlighted in our experiments, especially in conditional generation tasks. We also witnessed that DiffusionTS and ImagenTime demonstrated more cross-task versatility than classical GANs and VAEs, achieving competitive performances over synthesis, forecasting, and imputation. Consequently, they stand out as a robust and advisable choice for general-purpose time series generation. 

On the other hand, task-specific model selection is more nuanced and context-dependent. For time series synthesis, we recommend COSCIGAN and ImagenTime, respectively, for their strong ability to fit closely to the complex data distribution and generate useful time series for downstream tasks. FourierFlow is also recommended for univariate synthesis (see appendix \ref{chap:univar_exp}). TimeVQVAE is then advised for class label-guided synthesis. We advocate that users commence with our naive baselines as well, to explore more efficient generative time series models for class label-guided generation. For time series forecasting, we suggest beginning with CSDI, which could produce reasonable predictions in general cases. TMDM is also advisable for the non-stationary time series. For time series imputation, we highly recommend diffusion-based models, especially CSDI, as the first choice for their absolute advantages compared to others.

Besides, we noticed that differential equation-based models generally underperformed on our benchmarks but we believed that their inherent inductive bias suggests they are promising but under-explored for continuous time series generation.


\textbf{Potential Future Directions}. 1) \textit{Generative Time Series Foundation Models}: Recent models \cite{wang2025a, kollovieh2023predict, yuan2024diffusion, naiman2024utilizing, tashiro2021csdi} are increasingly designed for multiple time series tasks. A critical research direction is to investigate generative time series foundation models. We expect to develop foundation models in a novel generative paradigm, where they are trained to fit the global data distributions and inferred controllably for diverse downstream tasks.  2) \textit{Robust Evaluation}: Existing model-based metrics could have robustness issues \cite{ang2023tsgbench}, especially PS and DS, which are sensitive to the auxiliary evaluator models. Their variants, e.g., using a linear post-hoc regressor \cite{kollovieh2023predict}, may still limit their downstream application. Therefore, developing more robust and holistic evaluation metrics constitutes a critical research direction. 3) \textit{Online Development}: Current generative time series models are mostly operated under the static environments and could have large computation burdens (see our appendix \ref{chap:appendix_time}), especially Diffusions. Real-world time series, however, are dynamically changing, along with complex temporal distribution shifts. How to develop online adaptive paradigms with tolerable latency could be another practical research direction.

\section{Conclusion and future works}
In this paper, we thoroughly investigated both generative time series models and the related libraries, and revealed the lack of a comprehensive benchmark library for versatile generative time series models. Then, we introduced our developed \texttt{GenTS} library, including multi-domain time series datasets, multi-type SOTA generative models, and various evaluation metrics. Modularly designed, it can be easily extended and customized by users. Finally, based on the developed \texttt{GenTS}, we conducted comprehensive benchmarking experiments under different time series tasks. Accordingly, we gave discussions on model selection and future research directions. 

In the future, we will 1) continuously incorporate more high-quality time series datasets and the most recent promising generative models, 2) broaden our scope for more emerging generative time series analysis tasks, and 3) explore more robust and systematic evaluation frameworks for cross-task and task-specific contexts. 



\begin{acks}
    This work was supported in part by the National Natural Science Foundation of China (52477130), and in part by the Research Grants Council of the Hong Kong SAR (HKU 17200224). 	
\end{acks}

\bibliographystyle{ACM-Reference-Format}
\bibliography{sample-base}

\appendix
\section{\texttt{GenTS} Details}
In this appendix, we will complement some details of \texttt{GenTS}. We also provide a comprehensive document website for \texttt{GenTS}, including installation, tutorials on various cases (including how to customize datasets and models), and detailed API documents.
Please check \url{https://willwang1113.github.io/GenTS/} for more information.

\subsection{Simulation of Spiral2D and SineND}\label{apendix:data}
The \texttt{Spiral2D} dataset is introduced as a synthetic dataset. Each sample in the dataset is a two-dimensional time series representing a spiral curve. The coordinates for a given curve are generated using a parametric equation in polar coordinates, where the radius $r(t)$ is a linear function of the angle $t$: 
$$r(t) = a + bt,$$ 
with $a \sim \mathcal{U}[0, 0.5)$, $b \sim \mathcal{U}[0, 0.2)$. Specifically, for $t$ ranging from 0 to 4$\pi$, the coordinates $(x_{1}(t), x_{2}(t))$ are given by:$$x_1(t)= \pm r(t) \cos(t),$$ $$x_2(t) = r(t) \sin(t).$$The sign of the $x_{1}(t)$ is randomly chosen, resulting in two classes: clock-wise and counter clock-wise spirals. Therefore, \texttt{Spiral2D} is a simulated dataset with class labels. Small-variance Gaussian noise is added to each coordinate to ensure diversity. 

Likewise, the \texttt{SineND} dataset is a synthetic dataset. Each sample is an $N$-variate time series where each dimension represents an independent sine wave. The number of variates can be decided by users. The value for each variate at time $t$ is generated by the equation:$$x(t) = \sin(at+b),$$ where the frequency $a \sim \mathcal{U}[0.05, 0.4]$ and the phase $b \sim \mathcal{U}[0, 1.5]$ are drawn independently for each dimension of each sample. Diversity across the dataset is ensured by the parameter randomization, and the resulting waves are normalized to the range $[0, 1]$. 

\subsection{Vanilla Models}\label{chap:model}
In our \texttt{GenTS}, we include four naive benchmark models, VanillaVAE, VanillaGAN, VanillaDDPM, and VanillaMAF.

\textbf{VanillaVAE}: We build a naive VAE with Multilayer Perceptrons (MLP) as encoders and decoders. To integrate conditions like look-back windows, and the observed values, we also utilize MLP as a condition encoder. For class label conditions, we simply use a \texttt{nn.Embedding} layer as label encoder. The VanillaVAE is trained in the paradigm of $\beta$-VAE.

\textbf{VanillaGAN}: Similar to VanillaVAE, we construct VanillaGAN with MLP as both generator and discriminator. The condition encoder is also the same as VanillaVAE. The VanillaVAE is trained in the paradigm of Wasserstein GAN.

\textbf{VanillaDDPM}: For VanillaDDPM, we utilize  DiT \cite{peebles2023scalable} (in the 1D version) as the denoising backbone. The condition encoder is also the same as VanillaVAE, and adds the condition embedding to the diffusion step embedding. We also provide choices of noise schedule and predict $x_0$ or $\epsilon$ as arguments for training.

\textbf{VanillaMAF}: We follow \cite{germain2015made} to construct a naive masked autoregressive flow model. For the input time series in the shape of $[N, T, D]$, we simply flatten into $[N, T\times D]$ to fit in the model. The condition encoder is the same as the VanillaVAE.

\subsection{Evaluation Metrics}\label{chap:metrics}
\textbf{Discriminative Score} \cite{yoon2019time}: To calculate DS, we need to train a post-hoc GRU classifier. We mix the generated time series and the real time series, with labels of \textit{False} and \textit{True}. Then, we split the mixed dataset into $8:2$ as training and testing sets. The classifier is trained to tell apart whether the time series is real or not, and then tested on the hold-out set. The accuracy will be recorded. Ideally, if the generated data is highly similar to the real one, then the classifier should randomly guess whether the data sample is generated or not, i.e., the accuracy will be around 0.5. Therefore, DS is calculated to be as close to 0.5 as possible:
$$DS = \mid 0.5- acc \mid $$

\textbf{Predictive Score} \cite{yoon2019time}: For PS, we train a post-hoc GRU regressor. Trained on the synthetic time series, it performs one-step ahead time series forecasting. Then, the test error (Mean Absolute Error, MAE) of the trained RNN will be reported on the real data. This is also known as Train on Synthetic Test on Real (TSTR).

\textbf{Context-FID} \cite{jeha2022psa}: Adapting from the Frechet Inception Distance (FID), it evaluates the quality of synthetic data by measuring the alignment with the local temporal context. Leveraging embeddings from TS2Vec \cite{franceschi2019unsupervised}, C-FID calculats Frechet Distance between the latent representations of generated data and real data. For each time series dataset, TS2Vec will be retrained, thus evaluating the contextual similarity.

\section{Experiments}\label{chap:exp_appendix}
All of our experiments are launched on one NVIDIA GeForce RTX 4090 24GB GPU and four NVIDIA GeForce RTX 3080 Ti 12GB GPUs. All the training processes are unified within a maximum of 300 epochs with an early stopping mechanism (10 epochs patience).

\subsection{Time Series Synthesis}\label{chap:univar_exp}
\textbf{Univariate results} 
Table. \ref{tab:syn_cfid_uni}, \ref{tab:syn_wd_uni}, \ref{tab:syn_ps_uni}, \ref{tab:syn_ds_uni} records the univariate time series synthesis results. 

Similar to the multivariate results, COSCIGAN and ImagenTime still showed superior performances, respectively, on distributional fidelity and downstream usefulness. Besides, we noticed that FourierFlow demonstrated competitive and balanced performances in the univariate setting on all four metrics, becoming a reasonable model choice in this context.

Another univariate time series model, FIDE, showed limited advantages. Designed for capturing extreme values in the time series, FIDE only showed competitive performance over Stocks and ETTh, while it even lagged far behind other models in most cases. Some parallel research works found that it may be suitable for some vertical applications, like financial markets \cite{ang2025ctbench}.

\begin{table*}[htb]
    \centering
    \caption{Full results of Context-FID on the univariate time series synthesis task. The best model on each dataset is in bold; the second and third-best are underlined. The last column reports the frequency of a model ranking top-3.}
    \label{tab:syn_cfid_uni}
    \resizebox{\textwidth}{!}{
    \begin{tabular}{l|cccccccccccc|c}
\toprule
\textbf{Model} & AirQuality & ECG & ETTh1 & ETTh2 & Electricity & Energy & Exchange & MuJoCo & SineND & Stocks & Traffic & Weather & \textbf{\textit{Top3 count}} \\

\midrule
\textbf{COSCIGAN}         & {\ul 0.331}    & 0.066          & {\ul 0.078}    & {\ul 0.058}    & {\ul 0.107}       & {\ul 0.177}       & {\ul 0.146}       & {\ul 0.190}       & 1.291             & {\ul 0.171}    & 0.633          & {\ul 0.038}    & \textbf{9}                            \\
\textbf{DiffusionTS}      & 1.600          & 0.443          & 1.087          & 0.482          & 0.759             & 2.326             & 0.639             & 6.626             & 0.231             & 5.283          & 1.505          & 0.605          & 0                            \\
\textbf{FIDE}             & >100            & 14.334         & {\ul 0.050}    & 0.116          & \textgreater{}100 & \textgreater{}100 & \textgreater{}100 & 4.244             & \textgreater{}100 & {\ul 0.223}    & 22.427         & 0.075          & 2                            \\
\textbf{FourierDiffusion} & 0.373          & 0.529          & 0.298          & {\ul 0.052}    & 0.699             & 4.584             & 5.995             & 0.408             & 17.114            & 3.529          & 3.824          & {\ul 0.055}    & 2                            \\
\textbf{FourierFlow}      & 0.508          & \textbf{0.025} & \textbf{0.048} & \textbf{0.030} & {\ul 0.116}       & {\ul 0.105}       & 0.861             & 1.171             & 0.212             & 1.892          & {\ul 0.139}    & 10.003         & \textbf{6}                            \\
\textbf{GTGAN}            & 4.897          & 6.909          & 2.646          & 0.585          & 5.382             & 13.536            & 0.434             & 0.302             & 4.007             & 2.885          & 7.131          & 5.446          & 0                            \\
\textbf{ImagenTime}       & {\ul 0.354}    & {\ul 0.031}    & 0.146          & 0.115          & \textbf{0.014}    & \textbf{0.062}    & 0.540             & 0.291             & {\ul 0.095}       & \textbf{0.116} & \textbf{0.024} & 0.210          & \textbf{7}                            \\
\textbf{KoVAE}            & \textbf{0.283} & 0.423          & 1.049          & 1.234          & 0.345             & 0.663             & {\ul 0.141}       & \textbf{0.058}    & 1.192             & 2.479          & 1.049          & \textbf{0.012} & 4                            \\
\textbf{LS4}              & 4.897          & 1.303          & 2.301          & 3.651          & 1.468             & 8.615             & 2.365             & 2.785             & 2.781             & 4.117          & 5.019          & 0.951          & 0                            \\
\textbf{LatentODE}        & 4.897          & 6.351          & 2.992          & 3.472          & 6.984             & 5.421             & 4.780             & 2.863             & 16.547            & 2.608          & 2.987          & 5.465          & 0                            \\
\textbf{LatentSDE}        & 1.674          & 4.269          & 4.654          & 1.771          & 4.570             & 4.087             & 0.625             & 11.981            & \textgreater{}100 & 0.861          & 11.175         & 3.178          & 0                            \\
\textbf{PSAGAN}           & 0.656          & 1.114          & 1.244          & 2.233          & 2.298             & 1.942             & 4.542             & 1.969             & 0.918             & 3.310          & 3.642          & 2.373          & 0                            \\
\textbf{RCGAN}            & 0.564          & 5.551          & 1.855          & 1.726          & 0.235             & 4.606             & 52.276            & 21.085            & 83.819            & 0.459          & 2.653          & 5.366          & 0                            \\
\textbf{SDEGAN}           & 52.795         & 9.994          & 7.641          & 3.705          & 15.384            & 3.458             & 66.702            & \textgreater{}100 & \textgreater{}100 & 6.833          & 7.337          & 1.038          & 0                            \\
\textbf{TimeGAN}          & 4.597          & 0.455          & 0.618          & 0.860          & 4.951             & 1.126             & 0.776             & 1.903             & 0.501             & 0.809          & 5.337          & 2.233          & 0                            \\
\textbf{TimeVAE}          & 0.438          & 0.079          & 0.275          & 0.221          & 0.276             & 0.559             & 0.214             & {\ul 0.068}       & 0.380             & 0.311          & 0.613          & 0.100          & 1                            \\
\textbf{TimeVQVAE}        & 1.323          & 2.190          & 0.711          & 1.185          & 2.345             & 3.259             & 1.705             & 3.343             & 0.985             & 2.617          & 3.448          & 1.921          & 0                            \\
\textbf{VanillaDDPM}      & 2.838          & 0.235          & 0.233          & 1.265          & 0.247             & 0.245             & 0.830             & 1.929             & \textbf{0.083}    & 0.574          & {\ul 0.069}    & 3.725          & 2                            \\
\textbf{VanillaGAN}       & 0.758          & 1.316          & 5.195          & 3.437          & 4.189             & 2.778             & 0.515             & 5.329             & \textgreater{}100 & 2.682          & 2.680          & 1.762          & 0                            \\
\textbf{VanillaMAF}       & 0.518          & {\ul 0.039}    & 0.232          & 0.626          & 0.378             & 0.834             & \textbf{0.108}    & 0.219             & {\ul 0.208}       & 1.371          & 0.242          & 0.213          & 3                            \\
\textbf{VanillaVAE}       & 0.442          & 1.944          & 2.038          & 1.187          & 1.676             & 1.499             & 1.964             & 0.428             & 0.933             & 2.816          & 2.487          & 1.100          & 0                            \\
\bottomrule
\end{tabular}
}
\end{table*}

\begin{table*}[htb]
    \centering
    \caption{Full results of Wasserstein Distance  on the univariate time series synthesis task. The best model on each dataset is in bold; the second and third-best are underlined. The last column reports the frequency of a model ranking top3.}
    \label{tab:syn_wd_uni}
    \resizebox{\textwidth}{!}{
    \begin{tabular}{l|cccccccccccc|c}
\toprule
\textbf{Model} & AirQuality & ECG & ETTh1 & ETTh2 & Electricity & Energy & Exchange & MuJoCo & SineND & Stocks & Traffic & Weather & \textbf{\textit{Top3 count}} \\

\midrule
\textbf{COSCIGAN}         & \textbf{0.208} & {\ul 0.124}    & {\ul 0.134}    & 0.245          & {\ul 0.087}    & {\ul 0.140}       & {\ul 0.141}    & {\ul 0.061}    & 0.296          & {\ul 0.131}    & 0.433          & {\ul 0.101}    & \textbf{9}                            \\
\textbf{DiffusionTS}      & 0.289          & 0.385          & 0.423          & 0.484          & 0.468          & 0.523             & 0.309          & 0.337          & 0.162          & 0.515          & 0.482          & 0.474          & 0                            \\
\textbf{FIDE}             & >100            & 1.788          & {\ul 0.146}    & {\ul 0.235}    & 81.973         & \textgreater{}100 & 3.619          & 0.277          & 4.633          & {\ul 0.131}    & 3.062          & 0.151          & 3                            \\
\textbf{FourierDiffusion} & 0.393          & 0.317          & 0.209          & 0.241          & 0.151          & {\ul 0.167}       & 0.801          & 0.088          & 0.518          & 0.371          & {\ul 0.106}    & {\ul 0.111}    & 3                            \\
\textbf{FourierFlow}      & 0.333          & \textbf{0.066} & \textbf{0.091} & \textbf{0.146} & {\ul 0.104}    & \textbf{0.098}    & 0.303          & 0.126          & {\ul 0.075}    & 0.230          & {\ul 0.163}    & 1.751          & \textbf{7}                           \\
\textbf{GTGAN}            & 0.772          & 0.922          & 0.390          & 0.343          & 0.451          & 1.033             & 0.188          & 0.065          & 0.244          & 0.455          & 0.586          & 1.019          & 0                            \\
\textbf{ImagenTime}       & 0.309          & {\ul 0.101}    & 0.192          & {\ul 0.162}    & \textbf{0.068} & 0.211             & 0.315          & 0.082          & {\ul 0.072}    & \textbf{0.111} & \textbf{0.098} & 0.233          & \textbf{6}                            \\
\textbf{KoVAE}            & {\ul 0.284}    & 0.278          & 0.367          & 0.451          & 0.445          & 0.487             & {\ul 0.118}    & \textbf{0.029} & 0.146          & 0.373          & 0.530          & \textbf{0.058} & 4                            \\
LS4                       & 0.769          & 0.877          & 0.785          & 1.024          & 0.489          & 2.236             & 0.423          & 0.244          & 0.422          & 0.503          & 0.964          & 0.397          & 0                            \\
\textbf{LatentODE}        & 0.784          & 0.839          & 0.880          & 0.849          & 0.923          & 0.965             & 0.612          & 0.239          & 0.611          & 0.414          & 0.985          & 0.791          & 0                            \\
\textbf{LatentSDE}        & 0.595          & 0.785          & 0.589          & 0.632          & 0.786          & 0.838             & 0.268          & 0.421          & 3.149          & 0.211          & 0.630          & 0.629          & 0                            \\
\textbf{PSAGAN}           & 0.551          & 0.700          & 0.459          & 0.603          & 0.874          & 0.892             & 0.561          & 0.162          & 0.240          & 0.423          & 0.860          & 0.579          & 0                            \\
\textbf{RCGAN}            & 0.400          & 1.231          & 0.606          & 0.660          & 0.130          & 0.444             & 3.066          & 0.609          & 1.667          & 0.141          & 0.780          & 1.389          & 0                            \\
\textbf{SDEGAN}           & 0.778          & 2.581          & 0.844          & 1.147          & 0.565          & 0.941             & 1.906          & 1.758          & 1.977          & 0.442          & 0.600          & 0.550          & 0                            \\
\textbf{TimeGAN}          & 0.755          & 0.287          & 0.320          & 0.580          & 0.677          & 0.373             & 0.252          & 0.131          & 0.137          & 0.215          & 1.134          & 0.520          & 0                            \\
\textbf{TimeVAE}          & {\ul 0.284}    & 0.125          & 0.191          & 0.293          & 0.160          & 0.337             & 0.150          & {\ul 0.035}    & 0.090          & 0.158          & 0.359          & 0.161          & 2                            \\
\textbf{TimeVQVAE}        & 0.436          & 0.624          & 0.387          & 0.427          & 0.447          & 0.681             & 0.310          & 0.224          & 0.167          & 0.308          & 0.688          & 0.465          & 0                            \\
\textbf{VanillaDDPM}      & 0.488          & 0.300          & 0.185          & 0.537          & 0.223          & 0.232             & 0.445          & 0.208          & 0.099          & 0.161          & 0.187          & 1.000          & 0                            \\
\textbf{VanillaGAN}       & 0.296          & 0.765          & 0.862          & 0.847          & 0.915          & 1.057             & 0.341          & 0.335          & 2.473          & 0.405          & 1.047          & 0.671          & 0                            \\
\textbf{VanillaMAF}       & 0.308          & 0.126          & 0.206          & 0.427          & 0.197          & 0.224             & \textbf{0.107} & 0.066          & \textbf{0.068} & 0.297          & 0.257          & 0.217          & 2                            \\
\textbf{VanillaVAE}       & 0.402          & 0.483          & 0.506          & 0.520          & 0.422          & 0.534             & 0.415          & 0.082          & 0.204          & 0.341          & 0.605          & 0.437          & 0                            \\
\bottomrule
\end{tabular}
}
\end{table*}

\begin{table*}[htb]
    \centering
    \caption{Full results of Predictive Score on the univariate time series synthesis task. The best model on each dataset is in bold; the second and third-best are underlined. The last column reports the frequency of a model ranking top3.}
    \label{tab:syn_ps_uni}
    \resizebox{\textwidth}{!}{
    \begin{tabular}{l|cccccccccccc|c}
\toprule
\textbf{Model} & AirQuality & ECG & ETTh1 & ETTh2 & Electricity & Energy & Exchange & MuJoCo & SineND & Stocks & Traffic & Weather & \textbf{\textit{Top3 count}} \\

\midrule
\textbf{COSCIGAN}         & 0.169          & 0.226          & 0.258          & 0.271          & 0.245          & 0.308          & 0.059          & 0.015          & 0.282          & 0.055          & 0.219          & 0.024          & 0                            \\
\textbf{DiffusionTS}      & 0.170          & 0.225          & 0.229          & 0.235          & 0.251          & 0.289          & 0.027          & 0.020          & 0.038          & 0.017          & 0.194          & 0.012          & 0                            \\
\textbf{FIDE}             & 0.323          & 1.525          & 0.231          & 0.230          & 0.246          & 0.287          & 0.061          & 0.095          & 1.095          & 0.031          & 0.201          & 0.016          & 0                            \\
\textbf{FourierDiffusion} & 0.166          & 0.251          & 0.250          & 0.233          & 0.264          & 0.326          & 0.048          & 0.058          & 0.159          & 0.047          & 0.246          & 0.014          & 0                            \\
\textbf{FourierFlow}      & {\ul 0.163}    & {\ul 0.167}    & {\ul 0.217}    & {\ul 0.221}    & {\ul 0.230}    & {\ul 0.275}    & 0.026          & 0.009          & 0.029          & 0.018          & {\ul 0.162}    & \textbf{0.008} & \textbf{8}                            \\
\textbf{GTGAN}            & 0.448          & 0.430          & 0.278          & 0.374          & 0.352          & 0.363          & {\ul 0.026}    & 0.012          & 0.052          & 0.027          & 0.290          & 0.487          & 1                            \\
\textbf{ImagenTime}       & \textbf{0.162} & \textbf{0.162} & \textbf{0.213} & \textbf{0.219} & \textbf{0.218} & \textbf{0.270} & {\ul 0.025}    & 0.003          & {\ul 0.013}    & {\ul 0.016}    & \textbf{0.136} & {\ul 0.008}    & \textbf{11}                           \\
\textbf{KoVAE}            & 0.172          & 0.344          & 0.327          & 0.322          & 0.232          & 0.327          & 0.029          & 0.003          & 0.028          & 0.020          & 0.198          & {\ul 0.008}    & 1                            \\
LS4                       & 0.394          & 0.707          & 0.555          & 0.644          & 0.296          & 0.602          & 0.158          & 0.044          & 0.318          & 0.282          & 0.701          & 0.117          & 0                            \\
\textbf{LatentODE}        & 0.445          & 0.369          & 0.432          & 0.374          & 0.957          & 0.331          & 0.086          & 0.010          & \textbf{0.006} & 0.024          & 0.291          & 0.143          & 1                            \\
\textbf{LatentSDE}        & 0.178          & 0.408          & 0.251          & 0.247          & 0.247          & 0.297          & 0.026          & \textbf{0.002} & 0.019          & {\ul 0.016}    & 0.281          & 0.034          & 2                            \\
\textbf{PSAGAN}           & 0.228          & 0.868          & 0.456          & 0.461          & 0.399          & 0.414          & 0.192          & 0.104          & 0.435          & 0.056          & 0.556          & 0.040          & 0                            \\
\textbf{RCGAN}            & {\ul 0.162}    & 0.861          & 0.332          & 0.341          & 0.247          & 0.311          & 0.106          & 0.414          & 0.563          & 0.018          & 0.257          & 0.249          & 1                            \\
\textbf{SDEGAN}           & 0.172          & 0.442          & 0.258          & 0.246          & 0.254          & 0.308          & 0.032          & 0.068          & 0.146          & 0.029          & 0.275          & 0.018          & 0                            \\
\textbf{TimeGAN}          & 0.564          & 0.465          & 0.247          & 0.491          & 0.338          & 0.381          & 0.038          & 1.031          & 0.303          & 0.028          & 0.592          & 0.154          & 0                            \\
\textbf{TimeVAE}          & 0.188          & 0.179          & 0.255          & 0.275          & {\ul 0.225}    & {\ul 0.279}    & 0.027          & 0.008          & 0.035          & 0.021          & {\ul 0.165}    & 0.009          & 3                            \\
\textbf{TimeVQVAE}        & 0.253          & 0.365          & 0.304          & 0.382          & 0.302          & 0.391          & 0.137          & 0.033          & 0.105          & 0.104          & 0.295          & 0.086          & 0                            \\
\textbf{VanillaDDPM}      & 0.177          & 0.281          & 0.295          & 0.421          & 0.283          & 0.328          & \textbf{0.025} & 0.007          & 0.018          & \textbf{0.016} & 0.172          & 0.014          & 2                            \\
\textbf{VanillaGAN}       & 0.274          & 0.693          & 0.420          & 0.605          & 0.666          & 0.489          & 0.295          & 0.367          & 2.084          & 0.150          & 0.579          & 0.292          & 0                            \\
\textbf{VanillaMAF}       & 0.167          & {\ul 0.174}    & {\ul 0.223}    & {\ul 0.229}    & 0.238          & 0.287          & 0.026          & {\ul 0.002}    & {\ul 0.013}    & 0.020          & 0.170          & 0.009          & \textbf{5}                            \\
\textbf{VanillaVAE}       & 0.247          & 0.224          & 0.339          & 0.366          & 0.262          & 0.305          & 0.041          & {\ul 0.002}    & 0.019          & 0.031          & 0.238          & 0.020          & 1                            \\
\bottomrule
\end{tabular}
}
\end{table*}

\begin{table*}[htb]
    \centering
    \caption{Full results of Discriminative Score on the univariate time series synthesis task. The best model on each dataset is in bold; the second and third-best are underlined. The last column reports the frequency of a model ranking top3.}
    \label{tab:syn_ds_uni}
    \resizebox{\textwidth}{!}{
    \begin{tabular}{l|cccccccccccc|c}
\toprule
\textbf{Model} & AirQuality & ECG & ETTh1 & ETTh2 & Electricity & Energy & Exchange & MuJoCo & SineND & Stocks & Traffic & Weather & \textbf{\textit{Top3 count}} \\

\midrule
\textbf{COSCIGAN}         & 0.337          & 0.420          & 0.366          & 0.310          & 0.210          & 0.409          & 0.399          & 0.119          & 0.500          & 0.482          & {\ul 0.247}    & 0.437          & 1                            \\
\textbf{DiffusionTS}      & 0.297          & 0.451          & 0.313          & 0.304          & 0.344          & 0.401          & 0.315          & 0.367          & 0.489          & 0.308          & 0.296          & 0.360          & 0                            \\
\textbf{FIDE}             & 0.408          & 0.500          & {\ul 0.192}    & {\ul 0.178}    & {\ul 0.208}    & {\ul 0.186}    & 0.498          & 0.499          & 0.500          & 0.216          & 0.262          & 0.127          & 4                            \\
\textbf{FourierDiffusion} & \textbf{0.164} & 0.480          & 0.373          & 0.293          & 0.436          & 0.455          & 0.480          & 0.498          & 0.496          & 0.496          & 0.417          & {\ul 0.048}    & 2                            \\
\textbf{FourierFlow}      & 0.284          & \textbf{0.091} & {\ul 0.133}    & {\ul 0.067}    & {\ul 0.044}    & {\ul 0.133}    & {\ul 0.126}    & \textbf{0.089} & 0.314          & 0.221          & 0.270          & 0.363          & \textbf{7}                            \\
\textbf{GTGAN}            & 0.422          & 0.496          & 0.478          & 0.468          & 0.482          & 0.496          & 0.136          & 0.446          & 0.500          & 0.499          & 0.494          & 0.499          & 0                            \\
\textbf{ImagenTime}       & {\ul 0.178}    & {\ul 0.173}    & \textbf{0.099} & \textbf{0.061} & \textbf{0.029} & \textbf{0.036} & 0.144          & {\ul 0.101}    & {\ul 0.221}    & \textbf{0.059} & \textbf{0.057} & 0.068          & \textbf{10}                           \\
\textbf{KoVAE}            & 0.221          & 0.428          & 0.470          & 0.405          & 0.226          & 0.283          & 0.136          & 0.215          & 0.475          & 0.311          & 0.323          & \textbf{0.028} & 1                            \\
LS4                       & 0.422          & 0.494          & 0.496          & 0.493          & 0.458          & 0.454          & 0.489          & 0.483          & 0.500          & 0.500          & 0.500          & 0.494          & 0                            \\
\textbf{LatentODE}        & 0.421          & 0.499          & 0.490          & 0.499          & 0.500          & 0.499          & 0.471          & 0.492          & 0.489          & 0.391          & 0.495          & 0.491          & 0                            \\
\textbf{LatentSDE}        & 0.483          & 0.499          & 0.479          & 0.494          & 0.497          & 0.500          & 0.206          & 0.380          & 0.496          & 0.347          & 0.498          & 0.330          & 0                            \\
\textbf{PSAGAN}           & 0.442          & 0.497          & 0.495          & 0.499          & 0.494          & 0.470          & 0.447          & 0.458          & 0.500          & 0.467          & 0.500          & 0.398          & 0                            \\
\textbf{RCGAN}            & {\ul 0.198}    & 0.500          & 0.338          & 0.439          & 0.345          & 0.423          & 0.500          & 0.500          & 0.500          & {\ul 0.141}    & 0.456          & 0.468          & 2                            \\
\textbf{SDEGAN}           & 0.429          & 0.499          & 0.400          & 0.471          & 0.448          & 0.456          & 0.417          & 0.500          & 0.486          & 0.437          & 0.490          & 0.438          & 0                            \\
\textbf{TimeGAN}          & 0.490          & 0.490          & 0.403          & 0.479          & 0.473          & 0.434          & 0.215          & 0.429          & 0.493          & 0.210          & 0.500          & 0.441          & 0                            \\
\textbf{TimeVAE}          & 0.408          & 0.312          & 0.441          & 0.477          & 0.231          & 0.227          & \textbf{0.079} & 0.222          & 0.321          & {\ul 0.099}    & 0.346          & {\ul 0.045}    & 3                            \\
\textbf{TimeVQVAE}        & 0.394          & 0.499          & 0.374          & 0.487          & 0.499          & 0.479          & 0.478          & 0.419          & \textbf{0.179} & 0.478          & 0.481          & 0.487          & 1                            \\
\textbf{VanillaDDPM}      & 0.217          & 0.434          & 0.357          & 0.477          & 0.405          & 0.402          & 0.136          & 0.243          & 0.400          & 0.250          & 0.403          & 0.245          & 0                            \\
\textbf{VanillaGAN}       & 0.480          & 0.500          & 0.496          & 0.499          & 0.500          & 0.499          & 0.500          & 0.479          & 0.500          & 0.489          & 0.500          & 0.498          & 0                            \\
\textbf{VanillaMAF}       & 0.267          & {\ul 0.157}    & 0.195          & 0.200          & 0.273          & 0.238          & {\ul 0.119}    & {\ul 0.107}    & {\ul 0.279}    & 0.146          & {\ul 0.161}    & 0.154          & \textbf{5}                            \\
\textbf{VanillaVAE}       & 0.385          & 0.490          & 0.490          & 0.494          & 0.439          & 0.433          & 0.405          & 0.147          & 0.454          & 0.327          & 0.471          & 0.155          & 0                            \\
\bottomrule
\end{tabular}
}
\end{table*}

\textbf{Multivariate full results} Table.~\ref{tab:syn_cfid}, \ref{tab:syn_wd}, \ref{tab:syn_ps}, \ref{tab:syn_ds} recorded the full multivariate time series synthesis results, forming the Fig.~\ref{fig:syn_exp}. 

\begin{figure}[htb]
    \centering
    \includegraphics[width=0.95\linewidth]{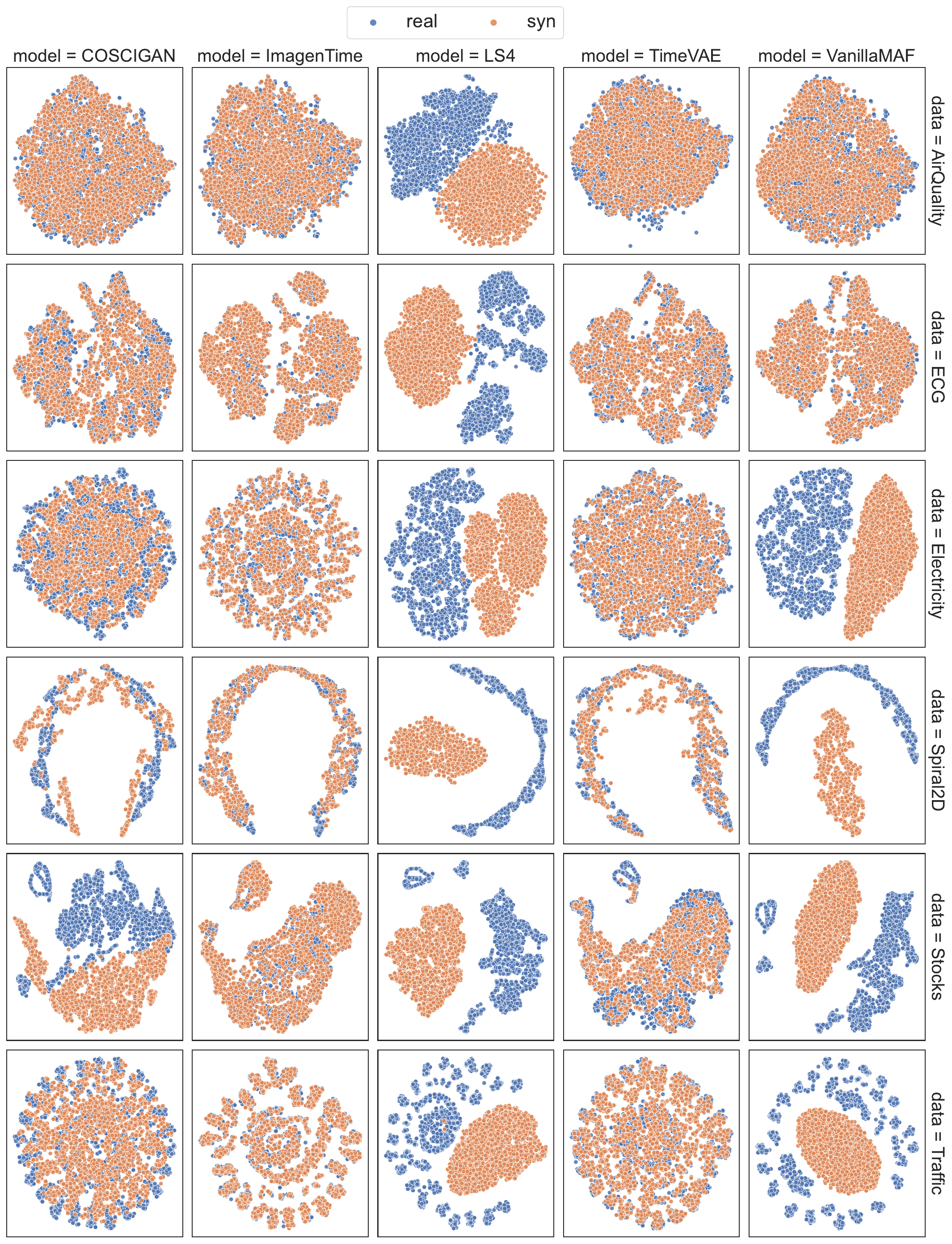}
    \caption{t-SNE Visualization of multivariate synthesis}
    \label{fig:tsne_multivar}
    \Description{The scatter plots shows the how well the representative models of each type fit on the different type of data distribution. The blue points are the real data, while the red points are the generated data.}
\end{figure}

Compared to the univariate setting, the advantage of COSCIGAN is more significant because of the special design of a central discriminator to coordinate multivariate correlation. Besides, the diffusion-based models still showed generally superior performances. We also provide a t-SNE visualization of representative models and datasets of each type, as shown in Figure~\ref{fig:tsne_multivar}. We can see that ImagenTime and TimeVAE showed coverage over the real time series data distribution, followed by COSCIGAN, indicating diversity and thoroughness.

\begin{table*}[htb]
    \centering
    \caption{Full results of Context-FID on the multivariate time series synthesis task. The best model on each dataset is in bold; the second and third-best are underlined. The last column reports the frequency of a model ranking top3.}
    \label{tab:syn_cfid}
    \resizebox{\textwidth}{!}{
    \begin{tabular}{l|ccccccccccccc|c}
\toprule
\textbf{Model} & AirQuality & ECG & ETTh1 & ETTh2 & Electricity & Energy & Exchange & MuJoCo & SineND & Spiral2D & Stocks & Traffic & Weather & \textbf{\textit{Top3 count}} \\
\midrule
\textbf{COSCIGAN}         & {\ul 0.178}    & \textbf{0.026} & \textbf{0.226} & {\ul 0.852}    & \textbf{0.445} & {\ul 0.257}    & \textbf{0.416} & \textbf{1.173} & 2.705          & {\ul 0.353}    & \textbf{0.692} & {\ul 0.810}    & 11.990         & 11         \\
\textbf{DiffusionTS}      & 0.677          & 0.535          & 0.916          & 1.308          & 2.986          & 1.894          & 1.529          & 7.337          & 1.424          & 1.749          & 1.642          & 1.588          & 12.520         & 0          \\
\textbf{FourierDiffusion} & \textbf{0.056} & 0.192          & 1.368          & 7.640          & 1.766          & \textbf{0.248} & 1.022          & >100        & >100        & 8.564          & >100        & 2.077          & {\ul 11.409}   & 3          \\
\textbf{GTGAN}            & 11.139         & 6.552          & 6.489          & 5.770          & 20.797         & 12.192         & 12.194         & 66.687         & 7.325          & 12.092         & 23.215         & 28.891         & 49.642         & 0          \\
\textbf{ImagenTime}       & {\ul 0.077}    & {\ul 0.048}    & 1.074          & 1.806          & {\ul 0.803}    & {\ul 0.857}    & 0.888          & {\ul 7.246}    & \textbf{0.374} & \textbf{0.144} & {\ul 0.926}    & \textbf{0.175} & {\ul 10.574}   & 10         \\
\textbf{KoVAE}            & 1.812          & 0.315          & 3.391          & 4.297          & 4.319          & 4.293          & 4.362          & 13.799         & 2.409          & 2.725          & 3.556          & 5.687          & 14.764         & 0          \\
\textbf{LS4}              & 11.140         & 1.370          & 16.220         & 19.190         & 7.080          & 10.417         & 10.824         & >100        & 14.956         & 16.017         & 10.241         & 4.646          & >100      & 0          \\
\textbf{LatentODE}        & 11.139         & 5.687          & 7.791          & 8.916          & 7.770          & 9.799          & 7.110          & 22.555         & 12.261         & 8.158          & 7.111          & 3.802          & 22.265         & 0          \\
\textbf{LatentSDE}        & 5.323          & 3.514          & 5.217          & 3.556          & 12.480         & 3.858          & 3.852          & 20.274         & >100       & 9.443          & 9.422          & 7.777          & 18.859         & 0          \\
\textbf{PSAGAN}           & 5.689          & 0.980          & 6.907          & 7.864          & 10.057         & 11.852         & 9.955          & 21.827         & 11.402         & 6.347          & 4.573          & 4.035          & 23.573         & 0          \\
\textbf{RCGAN}            & 1.388          & 6.284          & 4.367          & 1.898          & 4.220          & 11.004         & 30.537         & 41.695         & >100        & 10.115         & 11.693         & 7.764          & 13.449         & 0          \\
\textbf{SDEGAN}           & 6.623          & 9.896          & 19.284         & 8.632          & 18.196         & 17.348         & 60.741         & 22.363         & >100        & 19.428         & 23.683         & 27.418         & 21.760         & 0          \\
\textbf{TimeGAN}          & 12.752         & 7.920          & 6.802          & 9.480          & 20.876         & 13.192         & 8.711          & 25.644         & 5.038          & 2.682          & 4.190          & 24.130         & 24.230         & 0          \\
\textbf{TimeVAE}          & 0.439          & 0.080          & {\ul 0.621}    & \textbf{0.633} & {\ul 1.536}    & 0.890          & {\ul 0.423}    & 7.594          & 1.894          & {\ul 0.334}    & {\ul 0.991}    & 1.645          & 13.262         & 6          \\
\textbf{TimeVQVAE}        & 3.776          & 0.378          & 3.842          & 2.233          & 4.785          & 6.348          & 3.409          & 13.040         & 3.818          & 2.521          & 2.114          & 5.314          & 15.157         & 0          \\
\textbf{VanillaDDPM}      & 0.527          & 0.129          & 0.892          & 2.171          & 1.732          & 1.390          & 1.980          & {\ul 5.129}    & {\ul 0.714}    & 0.758          & 1.874          & {\ul 0.715}    & \textbf{8.744} & 4          \\
\textbf{VanillaGAN}       & 2.007          & 1.266          & 37.451         & 9.910          & 8.390          & 9.056          & 17.191         & 25.659         & 5.020          & 4.696          & 3.529          & 50.772         & 71.940         & 0          \\
\textbf{VanillaMAF}       & 0.470          & {\ul 0.030}    & {\ul 0.329}    & {\ul 0.767}    & >100        & 9.753          & {\ul 0.450}    & 58.540         & 15.007         & 6.788          & 5.455          & 13.479         & 20.286         & 4          \\
\textbf{VanillaVAE}       & 1.580          & 1.212          & 3.356          & 3.271          & 5.336          & 3.322          & 5.206          & 18.739         & {\ul 0.935}    & 4.196          & 3.480          & 6.544          & 15.843         & 1          \\
\bottomrule
\end{tabular}
}
\end{table*}

\begin{table*}[htb]
    \centering
    \caption{Full results of Wasserstein Distance on the multivariate time series synthesis task. The best model on each dataset is in bold; the second and third-best are underlined. The last column reports the frequency of a model ranking top3.}
    \label{tab:syn_wd}
    \resizebox{\textwidth}{!}{
    \begin{tabular}{l|ccccccccccccc|c}
\toprule
\textbf{Model} & AirQuality & ECG & ETTh1 & ETTh2 & Electricity & Energy & Exchange & MuJoCo & SineND & Spiral2D & Stocks & Traffic & Weather & \textbf{\textit{Top3 count}} \\
\midrule
\textbf{COSCIGAN}         & \textbf{0.182} & {\ul 0.072}    & \textbf{0.105} & {\ul 0.332}    & \textbf{0.151} & {\ul 0.096}    & \textbf{0.153} & \textbf{1.359} & 0.187          & \textbf{0.157} & \textbf{0.197} & 0.230          & {\ul 0.296}    & 11         \\
\textbf{DiffusionTS}      & 0.383          & 0.261          & 0.237          & 0.389          & 0.392          & 0.292          & 0.265          & {\ul 2.907}    & 0.160          & 0.328          & 0.271          & 0.314          & {\ul 0.328}    & 2          \\
\textbf{FourierDiffusion} & 0.245          & 0.220          & 0.312          & 1.192          & {\ul 0.211}    & \textbf{0.083} & 0.202          & 6.356          & 3.198          & 0.745          & 3.060          & {\ul 0.161}    & \textbf{0.200} & 4          \\
\textbf{GTGAN}            & 0.921          & 0.886          & 0.422          & 0.803          & 0.836          & 0.640          & 0.802          & 6.148          & 0.184          & 0.675          & 0.802          & 0.690          & 1.534          & 0          \\
\textbf{ImagenTime}       & {\ul 0.201}    & 0.170          & 0.328          & 0.457          & {\ul 0.182}    & 0.236          & 0.218          & 5.860          & \textbf{0.064} & {\ul 0.135}    & {\ul 0.250}    & \textbf{0.092} & 0.355          & 6          \\
\textbf{KoVAE}            & 0.499          & 0.219          & 0.401          & 0.591          & 0.357          & 0.506          & 0.517          & 4.280          & 0.149          & 0.319          & 0.451          & 0.433          & 0.516          & 0          \\
\textbf{LS4}              & 0.913          & 1.111          & 1.373          & 1.016          & 0.569          & 0.884          & 1.004          & 8.827          & 0.530          & 1.349          & 1.015          & 0.608          & 13.889         & 0          \\
\textbf{LatentODE}        & 0.902          & 0.866          & 1.027          & 0.956          & 0.886          & 0.828          & 0.740          & 6.956          & 0.547          & 1.036          & 0.587          & 0.795          & 0.964          & 0          \\
\textbf{LatentSDE}        & 0.455          & 0.831          & 0.415          & 0.515          & 0.576          & 0.313          & 0.495          & 7.237          & 3.588          & 0.759          & 0.583          & 0.322          & 0.528          & 0          \\
\textbf{PSAGAN}           & 0.699          & 0.755          & 0.628          & 0.798          & 0.973          & 0.836          & 0.808          & 6.925          & 0.475          & 0.524          & 0.527          & 0.712          & 0.941          & 0          \\
\textbf{RCGAN}            & 0.426          & 1.039          & 0.529          & 0.480          & 0.378          & 0.548          & 1.224          & 8.230          & 1.261          & 1.072          & 0.862          & 0.723          & 0.780          & 0          \\
\textbf{SDEGAN}           & 0.504          & 1.518          & 0.665          & 0.546          & 0.591          & 0.726          & 1.557          & 7.210          & 2.058          & 1.398          & 0.909          & 0.862          & 0.660          & 0          \\
\textbf{TimeGAN}          & 0.978          & 0.942          & 0.290          & 0.951          & 0.991          & 0.292          & 0.769          & 7.182          & {\ul 0.106}    & 0.304          & 0.314          & 0.963          & 0.978          & 1          \\
\textbf{TimeVAE}          & 0.303          & {\ul 0.133}    & {\ul 0.226}    & {\ul 0.293}    & 0.227          & {\ul 0.185}    & {\ul 0.179}    & 4.558          & 0.139          & {\ul 0.120}    & {\ul 0.246}    & 0.229          & 0.369          & 7          \\
\textbf{TimeVQVAE}        & 0.553          & 0.214          & 0.490          & 0.341          & 0.361          & 0.553          & 0.354          & 3.341          & 0.168          & 0.322          & 0.296          & 0.375          & 0.530          & 0          \\
\textbf{VanillaDDPM}      & 0.310          & 0.242          & 0.246          & 0.683          & 0.261          & 0.271          & 0.224          & {\ul 2.400}    & {\ul 0.090}    & 0.273          & 0.355          & {\ul 0.149}    & 0.367          & 3          \\
\textbf{VanillaGAN}       & 0.524          & 0.668          & 1.502          & 0.970          & 1.134          & 0.830          & 0.919          & 7.683          & 0.213          & 0.523          & 0.511          & 1.731          & 1.629          & 0          \\
\textbf{VanillaMAF}       & {\ul 0.226}    & \textbf{0.068} & {\ul 0.184}    & \textbf{0.289} & 1.611          & 0.921          & {\ul 0.178}    & 7.832          & 0.659          & 0.512          & 0.586          & 0.927          & 0.874          & 5          \\
\textbf{VanillaVAE}       & 0.501          & 0.409          & 0.518          & 0.570          & 0.477          & 0.497          & 0.602          & 7.145          & 0.122          & 0.411          & 0.441          & 0.551          & 0.590          & 0          \\
\bottomrule
\end{tabular}

}

\end{table*}

\begin{table*}[htb]
\centering
\caption{Full results of Predictive Score on the multivariate time series synthesis task. The best model on each dataset is in bold; the second and third-best are underlined. The last column reports the frequency of a model ranking top3.}
\label{tab:syn_ps}
\resizebox{\textwidth}{!}{
\begin{tabular}{l|ccccccccccccc|c}
\toprule
\textbf{Model} & AirQuality & ECG & ETTh1 & ETTh2 & Electricity & Energy & Exchange & MuJoCo & SineND & Spiral2D & Stocks & Traffic & Weather & \textbf{\textit{Top3 count}} \\
\midrule
\textbf{COSCIGAN}         & {\ul 0.449}    & 0.178          & {\ul 0.626}    & 0.585          & 0.584          & 0.594          & 0.458          & \textbf{4.382} & 0.246          & \textbf{0.417} & 0.214          & {\ul 0.400}    & 0.660          & 5          \\
\textbf{DiffusionTS}      & 0.452          & 0.259          & 0.628          & 0.586          & 0.585          & {\ul 0.593}    & 0.458          & {\ul 4.385}    & 0.215          & {\ul 0.418}    & 0.202          & {\ul 0.398}    & {\ul 0.658}    & 5          \\
\textbf{FourierDiffusion} & {\ul 0.449}    & 0.262          & {\ul 0.625}    & {\ul 0.578}    & 0.586          & 0.594          & 0.457          & 4.388          & 0.271          & 0.458          & 0.269          & 0.420          & \textbf{0.658} & 4          \\
\textbf{GTGAN}            & 0.598          & 0.451          & 0.640          & 0.620          & 0.638          & 0.599          & 0.480          & 4.412          & 0.217          & 0.426          & 0.199          & 0.435          & 0.661          & 0          \\
\textbf{ImagenTime}       & \textbf{0.448} & {\ul 0.163}    & \textbf{0.622} & \textbf{0.577} & \textbf{0.573} & \textbf{0.593} & \textbf{0.455} & 4.391          & \textbf{0.214} & {\ul 0.417}    & \textbf{0.192} & \textbf{0.394} & {\ul 0.658}    & 12         \\
\textbf{KoVAE}            & 0.474          & 0.310          & 0.649          & 0.615          & 0.646          & 0.599          & 0.464          & 4.400          & 0.221          & 0.427          & 0.277          & 0.426          & 0.660          & 0          \\
\textbf{LS4}              & 0.611          & 1.339          & 0.737          & 0.667          & 0.677          & 0.608          & 0.572          & 4.426          & 0.264          & 0.555          & 0.391          & 0.458          & 0.680          & 0          \\
\textbf{LatentODE}        & 0.591          & 0.399          & 0.692          & 0.651          & 0.652          & 0.636          & 0.540          & 4.423          & 0.272          & 0.551          & 0.264          & 0.486          & 0.693          & 0          \\
\textbf{LatentSDE}        & 0.462          & 0.410          & 0.642          & 0.586          & 0.638          & 0.600          & 0.480          & 4.436          & 0.247          & 0.443          & 0.328          & 0.432          & 0.691          & 0          \\
\textbf{PSAGAN}           & 0.498          & 0.890          & 0.717          & 0.900          & 0.675          & 0.642          & 0.544          & 4.455          & 0.263          & 0.514          & 0.312          & 0.545          & 0.788          & 0          \\
\textbf{RCGAN}            & 0.453          & 2.099          & 0.632          & 0.592          & 0.623          & 0.602          & 0.521          & 4.489          & 0.607          & 1.015          & 0.555          & 0.437          & 0.663          & 0          \\
\textbf{SDEGAN}           & 0.459          & 0.415          & 0.667          & 0.594          & 0.655          & 0.616          & 0.498          & 4.499          & 0.229          & 0.431          & 0.239          & 0.450          & 0.673          & 0          \\
\textbf{TimeGAN}          & 0.650          & 0.368          & 0.773          & 0.626          & 0.687          & 0.816          & 0.595          & 4.444          & 0.233          & 0.448          & 0.588          & 0.677          & 0.698          & 0          \\
\textbf{TimeVAE}          & 0.454          & {\ul 0.175}    & 0.628          & 0.585          & {\ul 0.583}    & {\ul 0.593}    & {\ul 0.456}    & 4.388          & 0.219          & 0.418          & {\ul 0.194}    & 0.405          & 0.658          & 5          \\
\textbf{TimeVQVAE}        & 0.503          & 0.241          & 0.655          & 0.600          & 1.204          & 0.621          & 0.460          & 4.411          & 0.219          & 0.429          & 0.216          & 0.567          & 0.663          & 0          \\
\textbf{VanillaDDPM}      & 0.452          & 0.230          & 0.629          & 0.600          & {\ul 0.584}    & 0.595          & 0.463          & {\ul 4.385}    & {\ul 0.214}    & 0.548          & {\ul 0.198}    & 0.403          & 0.659          & 4          \\
\textbf{VanillaGAN}       & 0.463          & 0.602          & 0.903          & 0.663          & 0.766          & 0.686          & 0.576          & 4.478          & 0.264          & 0.546          & 0.405          & 0.566          & 0.725          & 0          \\
\textbf{VanillaMAF}       & 0.452          & \textbf{0.159} & 0.626          & {\ul 0.579}    & 0.810          & 0.864          & {\ul 0.456}    & 4.423          & 0.323          & 0.598          & 0.493          & 0.681          & 0.802          & 3          \\
\textbf{VanillaVAE}       & 0.473          & 0.215          & 0.639          & 0.594          & 0.627          & 0.596          & 0.495          & 4.406          & {\ul 0.214}    & 0.438          & 0.211          & 0.440          & 0.661          & 1          \\

\bottomrule
\end{tabular}

}
\end{table*}

\begin{table*}[htb]
    \centering
    \caption{Full results of Discriminative Score on the multivariate time series synthesis task. The best model on each dataset is in bold; the second and third-best are underlined. The last column reports the frequency of a model ranking top3.}
    \label{tab:syn_ds}
    \resizebox{\textwidth}{!}{

\begin{tabular}{l|ccccccccccccc|c}
\toprule
\textbf{Model} & AirQuality & ECG & ETTh1 & ETTh2 & Electricity & Energy & Exchange & MuJoCo & SineND & Spiral2D & Stocks & Traffic & Weather & \textbf{\textit{Top3 count}} \\
\midrule
\textbf{COSCIGAN}         & {\ul 0.073}    & {\ul 0.224}    & 0.470          & 0.431          & {\ul 0.473}    & 0.478          & 0.348          & \textbf{0.294} & 0.500          & {\ul 0.082}    & 0.437          & {\ul 0.425}    & 0.490          & 6          \\
\textbf{DiffusionTS}      & 0.292          & 0.474          & 0.355          & 0.409          & 0.496          & {\ul 0.426}    & 0.384          & {\ul 0.419}    & 0.354          & 0.261          & {\ul 0.329}    & {\ul 0.477}    & \textbf{0.364} & 5          \\
\textbf{FourierDiffusion} & \textbf{0.013} & 0.486          & 0.418          & {\ul 0.329}    & {\ul 0.489}    & 0.466          & 0.434          & 0.493          & 0.486          & 0.200          & 0.498          & 0.498          & {\ul 0.407}    & 4          \\
\textbf{GTGAN}            & 0.465          & 0.497          & 0.468          & 0.500          & 0.500          & 0.497          & 0.500          & 0.496          & 0.382          & 0.264          & 0.490          & 0.500          & 0.485          & 0          \\
\textbf{ImagenTime}       & {\ul 0.056}    & {\ul 0.115}    & \textbf{0.200} & \textbf{0.195} & \textbf{0.241} & \textbf{0.237} & \textbf{0.175} & 0.493          & \textbf{0.146} & {\ul 0.089}    & \textbf{0.164} & \textbf{0.316} & 0.449          & 11         \\
\textbf{KoVAE}            & 0.361          & 0.424          & 0.484          & 0.460          & 0.500          & 0.497          & 0.479          & 0.484          & 0.336          & 0.361          & 0.429          & 0.499          & 0.498          & 0          \\
\textbf{LS4}              & 0.464          & 0.499          & 0.500          & 0.500          & 0.500          & 0.500          & 0.497          & 0.500          & 0.500          & 0.500          & 0.500          & 0.500          & 0.500          & 0          \\
\textbf{LatentODE}        & 0.466          & 0.499          & 0.497          & 0.500          & 0.500          & 0.500          & 0.498          & 0.494          & 0.500          & 0.496          & 0.496          & 0.498          & 0.500          & 0          \\
\textbf{LatentSDE}        & 0.382          & 0.494          & 0.486          & 0.452          & 0.500          & 0.494          & 0.454          & 0.491          & 0.493          & 0.375          & 0.491          & 0.499          & 0.500          & 0          \\
\textbf{PSAGAN}           & 0.496          & 0.497          & 0.499          & 0.500          & 0.500          & 0.500          & 0.500          & 0.500          & 0.500          & 0.496          & 0.496          & 0.500          & 0.500          & 0          \\
\textbf{RCGAN}            & 0.329          & 0.500          & 0.499          & 0.497          & 0.499          & 0.500          & 0.500          & 0.500          & 0.500          & 0.500          & 0.500          & 0.500          & 0.465          & 0          \\
\textbf{SDEGAN}           & 0.485          & 0.494          & 0.499          & 0.496          & 0.500          & 0.500          & 0.500          & 0.500          & 0.496          & 0.454          & 0.497          & 0.500          & 0.500          & 0          \\
\textbf{TimeGAN}          & 0.499          & 0.499          & 0.499          & 0.500          & 0.500          & 0.500          & 0.500          & 0.500          & 0.468          & 0.321          & 0.499          & 0.500          & 0.500          & 0          \\
\textbf{TimeVAE}          & 0.339          & 0.274          & 0.385          & 0.443          & 0.492          & 0.454          & {\ul 0.314}    & {\ul 0.429}    & {\ul 0.250}    & \textbf{0.079} & 0.445          & 0.495          & 0.495          & 4          \\
\textbf{TimeVQVAE}        & 0.479          & 0.400          & 0.460          & 0.482          & 0.499          & 0.500          & 0.492          & 0.497          & 0.421          & 0.275          & {\ul 0.384}    & 0.497          & 0.490          & 1          \\
\textbf{VanillaDDPM}      & 0.259          & 0.419          & {\ul 0.353}    & 0.425          & 0.492          & {\ul 0.453}    & 0.438          & 0.440          & 0.350          & 0.232          & 0.406          & 0.479          & {\ul 0.433}    & 3          \\
\textbf{VanillaGAN}       & 0.466          & 0.499          & 0.500          & 0.499          & 0.500          & 0.500          & 0.499          & 0.499          & 0.486          & 0.257          & 0.500          & 0.500          & 0.500          & 0          \\
\textbf{VanillaMAF}       & 0.196          & \textbf{0.114} & {\ul 0.232}    & {\ul 0.346}    & 0.500          & 0.500          & {\ul 0.277}    & 0.500          & 0.500          & 0.500          & 0.499          & 0.500          & 0.500          & 4          \\
\textbf{VanillaVAE}       & 0.353          & 0.472          & 0.453          & 0.470          & 0.493          & 0.489          & 0.497          & 0.491          & {\ul 0.318}    & 0.304          & 0.426          & 0.495          & 0.491          & 1          \\
\bottomrule
\end{tabular}

}
\end{table*}

\subsection{Training and inference time}\label{chap:appendix_time}
We measured the training and inference overheads for all models across different tasks. 

For training, we measured the average time of 10 training steps of a batch of 64 time series samples. Similarly, sampling time was measured by averaging the sampling time at a batch size of 64 over 10 times. 

\begin{figure}[htb]
    \centering
    \includegraphics[width=\linewidth]{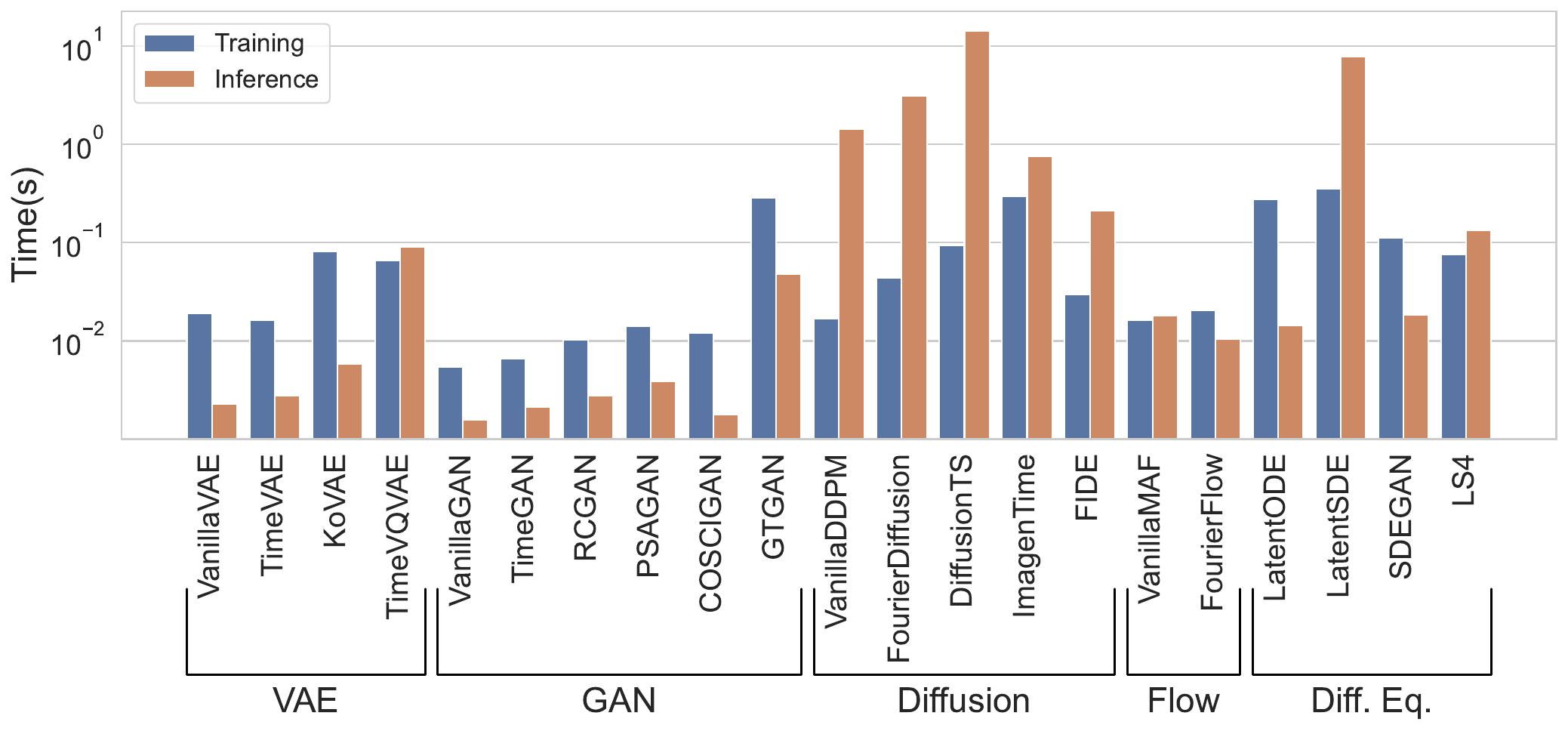}
    \caption{Training and inference time of different models in the univariate time series synthesis task}
    \Description{The grouped bar plot shows training and inference time of different models on time series synthesis.}
    \label{fig:syn_time}
\end{figure}

\begin{figure}[htb]
    \centering
    \includegraphics[width=\linewidth]{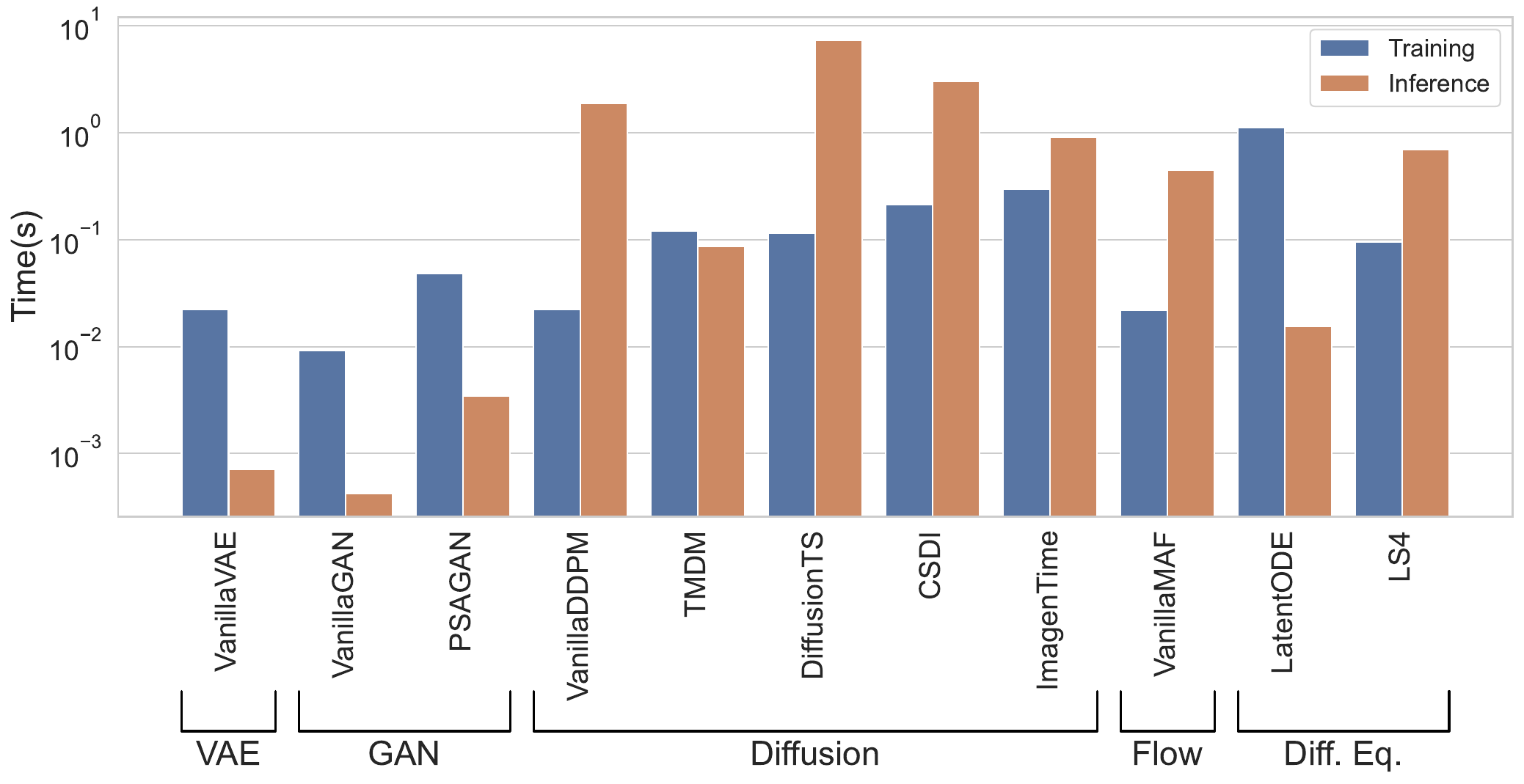}
    \caption{Training and inference time of different models in the time series forecasting task}
    \Description{The grouped bar plot shows training and inference time of different models on time series forecasting.}
    \label{fig:fcst_time}
\end{figure}

\begin{figure}[htb]
    \centering
    \includegraphics[width=\linewidth]{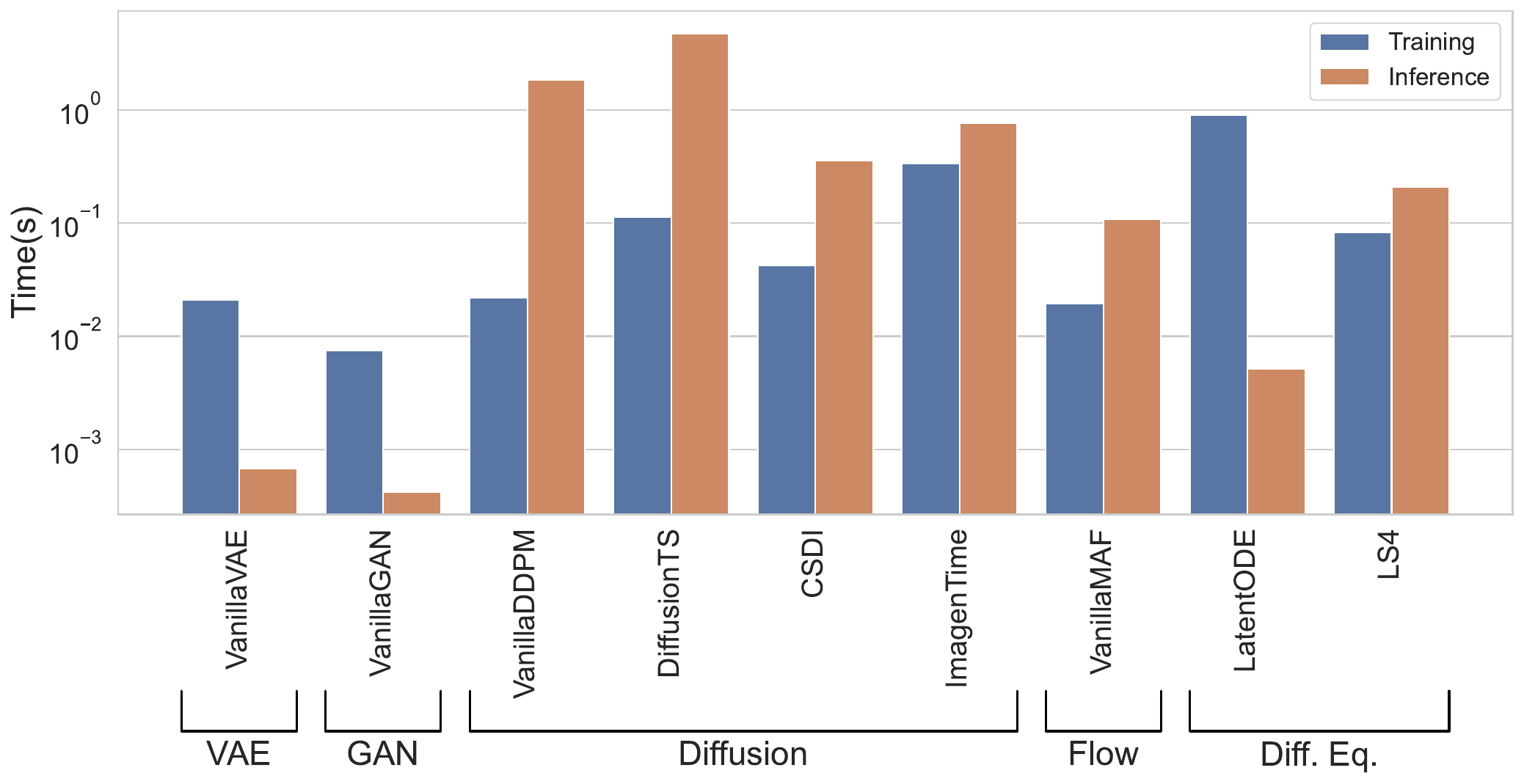}
    \caption{Training and inference time of different models in the time series imputation task}
    \Description{The grouped bar plot shows training and inference time of different models on time series imputation.}
    \label{fig:impute_time}
\end{figure}

As Figure~\ref{fig:syn_time}, \ref{fig:fcst_time}, and \ref{fig:impute_time} show, different kinds of models demonstrated distinct training and inference overheads. Most of the computation burden of GANs and VAEs lies in training, while their inference process is much faster than others. Diffusion-based models, designed with an iterative denoising process, generally have larger computation overheads on the inference stage, but their training speed is not significantly decreased, thus suitable for offline deployment. On the other hand, differential equation-based models, typically trained via frequently solving initial value problems, are naturally burdened with larger computation graphs for backwards.

\end{document}